\documentclass{article}

\usepackage{microtype}
\usepackage{graphicx}
\usepackage{subcaption}
\usepackage{booktabs} 
\usepackage{booktabs}
\usepackage{caption}
\usepackage[dvipsnames,table]{xcolor}
\usepackage{float} 
\usepackage{makecell}
\usepackage{seqsplit}

\usepackage{hyperref}

\usepackage[accepted]{icml2026}

\usepackage{amsmath}
\usepackage{amssymb}
\usepackage{mathtools}
\usepackage{amsthm}

\usepackage[capitalize,noabbrev]{cleveref}

\theoremstyle{plain}

\theoremstyle{definition}

\theoremstyle{remark}

\usepackage[textsize=tiny]{todonotes}
\usepackage{xspace}

\icmltitlerunning{Revisiting Pre-Propagation GNNs: Robust Diffusion Operators and Hidden-State Re-Propagation}

\begin{document}

\twocolumn[
  \icmltitle{Revisiting Pre-Propagation GNNs: Robust Diffusion Operators and Hidden-State Re-Propagation}



  \icmlsetsymbol{equal}{*}

  \begin{icmlauthorlist}
    \icmlauthor{Zichao Yue}{cornell}
    \icmlauthor{Zhiru Zhang}{cornell}
  \end{icmlauthorlist}

  \icmlaffiliation{cornell}{School of Electrical and Computer Engineering, Cornell University, Ithaca, New York, USA}

  \icmlcorrespondingauthor{Zichao Yue}{zy383@cornell.edu}

  \icmlkeywords{Machine Learning, ICML}

  \vskip 0.3in
]



\printAffiliationsAndNotice{}  
\newcommand{\PPG}{PP-GNN\xspace}
\newcommand{\PPGs}{PP-GNNs\xspace}
\newcommand{\MPG}{MP-GNN\xspace}
\newcommand{\MPGs}{MP-GNNs\xspace}
\begin{abstract}
  Pre-propagation graph neural networks (\PPGs) decouple node feature propagation from transformation: graph diffusion is performed once as preprocessing, and training reduces to dense per-node transformations. This design enables mini-batch training without inter-node dependencies, avoids repeated sparse matrix--matrix multiplications, and better matches modern accelerators optimized for dense compute. However, their expressivity remains unclear, and empirical results show a gap between \PPGs and their message-passing counterparts on commonly used graph benchmarks, especially heterophilic ones. In this paper, we propose a suite of robust graph diffusion operators for preprocessing and a few-shot hidden-state re-propagation scheme during training. Our methods improve the validation and test accuracy of \PPGs, enabling them to match the accuracy of message-passing GNNs while maintaining training efficiency.
\end{abstract}
\section{Introduction}
The message-passing scheme \cite{gilmer2017mpgnn} underlies most graph neural networks (GNNs), but its memory and computation costs grow exponentially with neighborhood expansion \cite{hamilton2017sage}, making large-scale training challenging. Prepropagation GNNs (\PPGs) have emerged as a scalable alternative \cite{wu2019SGC, frasca2020sign, dong2021pta, zhang2022gamlp, liao2022scara, chen2020gbp, deng2024hoga, zhu2020s2gc} by decoupling feature propagation from transformation. In \PPGs, node features are diffused over the graph once as a preprocessing step, and learning proceeds on the resulting diffused features via dense per-node transformations. This decoupling eliminates inter-node dependencies during training, making mini-batch optimization straightforward, avoiding repeated sparse matrix--matrix multiplications (SpMM), and better matching modern accelerators optimized for dense compute, while substantially improving efficiency and scalability compared to message-passing GNNs (\MPGs) \cite{yue2025graph}.

\begin{table}[t]
  \centering
  \caption{Motivating accuracy gap between \MPGs and \PPGs on heterophilic benchmarks. Entries report mean test accuracy on \texttt{roman-empire} and ROC AUC on \texttt{minesweeper}.}
  \small
  \setlength{\tabcolsep}{5pt}
  \renewcommand{\arraystretch}{0.95}
  \scalebox{0.92}{
  \begin{tabular}{lcc}
  \toprule
  \textbf{Model} & \textbf{Minesweeper} & \textbf{Roman-Empire} \\
  \midrule
  \multicolumn{3}{l}{\textbf{\MPGs}} \\
  SAGE & 93.51 $\pm$ 0.57 & 85.74 $\pm$ 0.67 \\
  GAT  & 93.91 $\pm$ 0.35 & 88.75 $\pm$ 0.41 \\
  DIR-GNN  & 87.05 $\pm$ 0.69 & 91.23 $\pm$ 0.32 \\
  \midrule
  \multicolumn{3}{l}{\textbf{\PPGs}} \\
  SIGN  & 90.71 $\pm$ 0.56 & 80.01 $\pm$ 0.50 \\
  HOGA  & 90.53 $\pm$ 0.66 & 79.39 $\pm$ 0.56 \\
  GAMLP & 90.47 $\pm$ 0.66 & 78.87 $\pm$ 0.65 \\
  \midrule
  \textbf{Best MP -- Best PP gap} &
  \textbf{3.20} & \textbf{11.22} \\
  \bottomrule
  \end{tabular}
  }
  \label{tab:motivation_gap}
  \vspace{-15pt}
  \end{table}
 \PPGs show competitive performance on large graph benchmarks (e.g., OGB), which are predominantly homophilic graphs. However, recent theoretical study suggests the scalability advantage of \PPGs may come with a sacrifice in expressivity \cite{chen2020graphaugmentedmlp}. Empirically, we find this limitation is most pronounced on heterophilic graphs: as shown in Table~\ref{tab:motivation_gap}, the test accuracy gap between \MPGs and \PPGs can reach up to 11\% on several heterophilic benchmarks. We hypothesize that two design aspects of \PPGs contribute to this gap: (i) diffusion operator limitations and (ii) one-shot feature propagation, and address them with more robust diffusion operators and hidden-state re-propagation (HRP).

 \textbf{First, diffusion-operator limitations.}
 A central design choice in \PPGs is the \emph{preprocessing diffusion operator} that generates multi-hop features.
Most existing \PPGs adopt simple operators---e.g., normalized adjacency or random-walk variants---whose spectral responses behave largely as \emph{low-pass} graph filters \cite{nt2019lowpass, gasteiger2019diffusion},
which can be suboptimal on heterophilic graphs that benefit from higher-frequency or band-pass components \cite{chien2020gprgnn, bo2021beyond}.
In principle, combining hop features from multi-step diffusion (e.g., $\{\Phi^k X\}_{k=0}^K$) enables richer spectral shaping; in practice, the hop budget $K$ is typically kept small due to efficiency considerations and the  \emph{oversmoothing} issue \cite{li2018oversmooth}.
Under such small $K$, the standard \emph{monomial} diffusion basis induced by powers of $\Phi$ can be poorly conditioned and yield highly correlated hop features, limiting approximation quality and destabilizing the learned hop combination.
To address this, we construct better-conditioned diffusion bases using \emph{orthogonal-polynomial} recurrences and \emph{Lanczos/Krylov} subspace methods (Jacobi-polynomial and Krylov-subspace bases), which reduce approximation error under the same hop budget and improve the stability of hop aggregation \cite{liao2019lanczosnet,wang2022powerful}.

 \textbf{Second, one-shot propagation}. Although \PPGs incorporate multi-hop information in preprocessing, this propagation is performed only once on the raw inputs and is not coupled to evolving hidden representations. Inspired by polynomial-filter GNNs like ChebNet\cite{defferrard2016convolutional}, where depth still increases expressivity even with large receptive fields per layer, we introduce hidden-state re-propagation to \PPGs, effectively adding additional rounds of graph filtering on task-adapted hidden states. A related idea is to reuse model predictions by treating logits---the learned soft labels---as additional node features \cite{wang2021bag, zhang2022gamlp}. However, we empirically show that replacing hidden states with logits in the re-propagation step leads to a substantial accuracy drop, suggesting that re-propagated hidden representations carry complementary information beyond class probabilities. In practice, only a few hidden-state re-propagation rounds are needed for convergence, preserving the training efficiency and scalability of vanilla \PPGs.
 
 To summarize, we study how to close the accuracy gap of \PPGs and \MPGs, especially on heterophilic graphs, while preserving their training efficiency and scalability. Our approach improves \PPGs by (i) using more robust diffusion operators and (ii) lightweight recoupling of feature propagation with representation learning. While motivated by heterophily, we observe that these improvements also transfer to homophilic benchmarks, yielding gains on a majority of them. Our main contributions are as follows:
\begin{itemize}
\setlength{\itemsep}{0pt plus 0.5pt}
\setlength{\parsep}{0pt}
\item We adapt Jacobi-polynomial and Krylov-subspace diffusion bases to the preprocessing regime of \PPGs, improving the conditioning of precomputed hop features and enabling richer spectral responses under a fixed hop budget.
\item We introduce hidden-state re-propagation, a mechanism that recouples feature propagation with learned representations and outperforms label-based reuse.
\item Across $4$ \PPGs and $12$ datasets, our method boosts test accuracy by $\mathbf{+2.18}$ avg on heterophilic ones and $\mathbf{+1.96}$ avg on homophilic ones, outperforming \MPG baselines on $\mathbf{7}$ out of 12 datasets.
\item Additionally, we explore an RNN-based hop aggregator as an efficient alternative to multi-head-attention-based aggregation, achieving comparable accuracy while significantly reducing end-to-end training time. 
\end{itemize}
\section{Preliminaries}
\label{sec:preliminaries}

\subsection{Graph Notation}
\label{sec:prelim-notation}
Let $G=(V,E)$ be a graph with $|V|=n$ nodes and adjacency matrix $A\in\mathbb{R}^{n\times n}$.
Let $X\in\mathbb{R}^{n\times d}$ denote input node features and $Y\in\{1,\dots,C\}^n$ node labels for a $C$-class node classification task.
We use $\mathcal{V}_{\text{tr}}, \mathcal{V}_{\text{va}}, \mathcal{V}_{\text{te}}$ to denote train/val/test node splits.
Let $D=\mathrm{diag}(d_1,\dots,d_n)$ be the degree matrix with $d_i=\sum_j A_{ij}$.
For a node $i$, $\mathcal{N}(i)$ denotes its neighbor set.
We write $\sigma(\cdot)$ for a pointwise nonlinearity and $\|\cdot\|$ for concatenation.

\subsection{Graph Spectral Theory and Spectral Graph Filters}
\label{sec:prelim-spectral}
For an undirected graph, the normalized Laplacian is
\begin{equation}
L \;=\; I - D^{-1/2} A D^{-1/2}.
\label{eq:norm-laplacian}
\end{equation}
Since $L$ is real symmetric, it admits an eigendecomposition
\begin{equation}
L \;=\; U \Lambda U^\top,
\label{eq:eig}
\end{equation}
where $U=[u_1,\dots,u_n]$ is orthonormal and $\Lambda=\mathrm{diag}(\lambda_1,\dots,\lambda_n)$ with $0=\lambda_1\le \cdots \le \lambda_n \le 2$.
Given a graph signal $x\in\mathbb{R}^n$ (e.g., one feature channel of $X$), its graph Fourier transform is $\hat{x}=U^\top x$ and inverse is $x=U\hat{x}$.

A \emph{spectral graph filter} is defined by a scalar response function $g:\mathbb{R}\to\mathbb{R}$ applied to the Laplacian spectrum:
\begin{equation}
g(L)x \;=\; U\,g(\Lambda)\,U^\top x,
\quad
g(\Lambda)=\mathrm{diag}\big(g(\lambda_1),\dots,g(\lambda_n)\big).
\label{eq:spectral-filter}
\end{equation}
Intuitively, smaller eigenvalues correspond to smoother (low-frequency) components, while larger eigenvalues correspond to more oscillatory (high-frequency) components; $g(\lambda)$ specifies how each frequency band is attenuated.

In practice, explicit eigendecomposition can be avoided by approximating the matrix function $g(L)$ directly as an operator on $L$, rather than evaluating $g$ on individual eigenvalues. A common choice is a \textbf{polynomial filter}, which approximates $g(L)$ by a degree-$K$ polynomial in $L$:
\begin{equation}
g(L)\;\approx\;\sum_{k=0}^{K}\alpha_k\,p_k(L),
\label{eq:poly-filter}
\end{equation}

\subsection{From Spectral Filters to Message-Passing GNNs}
\label{sec:prelim-mpg}
Spectral GNNs can be viewed as learning the filter $g(L)$ in~\eqref{eq:spectral-filter}. 
ChebNet-style models approximate $g(L)$ with truncated Chebyshev expansions as in~\eqref{eq:poly-filter}, where $p_k(\cdot)$ is the Chebyshev polynomial of degree $k$ and the coefficients $\{\alpha_k\}$ are learned.

GCN \cite{kipf2016gcn} can be derived as a particular low-order ($K{=}1$) approximation of ChebNet, resulting in a degree-1 polynomial filter in the Laplacian $L$. Consequently, each GCN layer performs essentially one-hop neighborhood mixing; larger multi-hop receptive fields are obtained by stacking layers, yielding a propagation rule based on a normalized adjacency-like operator.

Abstractly, many GNNs implement repeated local aggregation and transformation, which can be written in the message-passing (MP) form
\begin{equation}
\label{eq:mpnn}
h_i^{(\ell+1)}
=\phi^{(\ell)}\!\Big(
h_i^{(\ell)},\;
\mathrm{AGG}\big(\{\psi^{(\ell)}(h_i^{(\ell)},h_j^{(\ell)},e_{ij})\}\big)
\Big).
\end{equation}
\noindent
where $j\in\mathcal{N}(i)$, $h_i^{(\ell)}$ is the hidden state at layer $\ell$, $e_{ij}$ denotes optional edge features, and $\mathrm{AGG}$ is a permutation-invariant aggregator.
From this perspective, \MPGs can be interpreted as applying a graph filter to hidden representations interleaved with nonlinear transformations, bridging the spectral view and the spatial view.

\subsection{Pre-Propagation GNNs and Graph Diffusion Operators}
\label{sec:prelim-ppgnn}
Pre-propagation GNNs (\PPGs) decouple propagation from learnable transformation.
They first construct a \emph{graph diffusion operator} $\Phi\in\mathbb{R}^{n\times n}$ from topology, and apply it to node features before training a dense predictor.
Typical choices of $\Phi$ include normalized adjacency and random-walk operators.

Importantly, diffusion operators and spectral filters are closely related:
when $\Phi$ can be written as a function of the Laplacian, i.e.,
\begin{equation}
\Phi \;=\; g(L),
\label{eq:phi_as_filter}
\end{equation}
then applying diffusion $\Phi X$ is exactly applying a spectral graph filter with response $g(\lambda)$ as in~\eqref{eq:spectral-filter}.

Given $\Phi$, \PPGs precompute propagated features via multi-step diffusion operations. Multi-hop features are then obtained as
\begin{equation}
Z \;=\; \big(X,\ \Phi X,\ \Phi^2 X,\ \dots,\ \Phi^K X\big)\in\mathbb{R}^{n\times (K+1)d},
\label{eq:multihop-stack}
\end{equation}
which is then fed to a dense model $f_\theta$ (e.g., an MLP) to produce predictions:
\begin{equation}
H \;=\; f_\theta(Z),\qquad \hat{Y}=\mathrm{softmax}(H).
\label{eq:dense-predictor}
\end{equation}
\PPGs can be viewed as learning an implicit graph filter parameterized by $f_\theta$ over the subspace spanned by the diffusion powers of $\Phi$.
Since $f_\theta$ does not perform message passing, training can be implemented with dense kernels and mini-batching over nodes, often improving scalability compared to \MPGs.

\subsection{Label Usage as Input Features}
\label{sec:prelim-label}

Some methods augment node features with training label information.
Specifically, we construct a label-feature matrix $L_{\mathrm{in}}\in\mathbb{R}^{N\times C}$ whose $i$-th row is the one-hot encoding of $Y_i$ if node $i\in\mathcal{V}_{\mathrm{tr}}$, and the zero vector otherwise.
These label features (or \emph{soft labels} such as logits/probabilities from a previous stage) can be concatenated with node features and optionally diffused using the same operator, e.g., $\Phi L_{\mathrm{in}}$.

\section{Method}
\label{sec:method}
\begin{figure*}[t]
  \centering
  \includegraphics[width=0.8\textwidth]{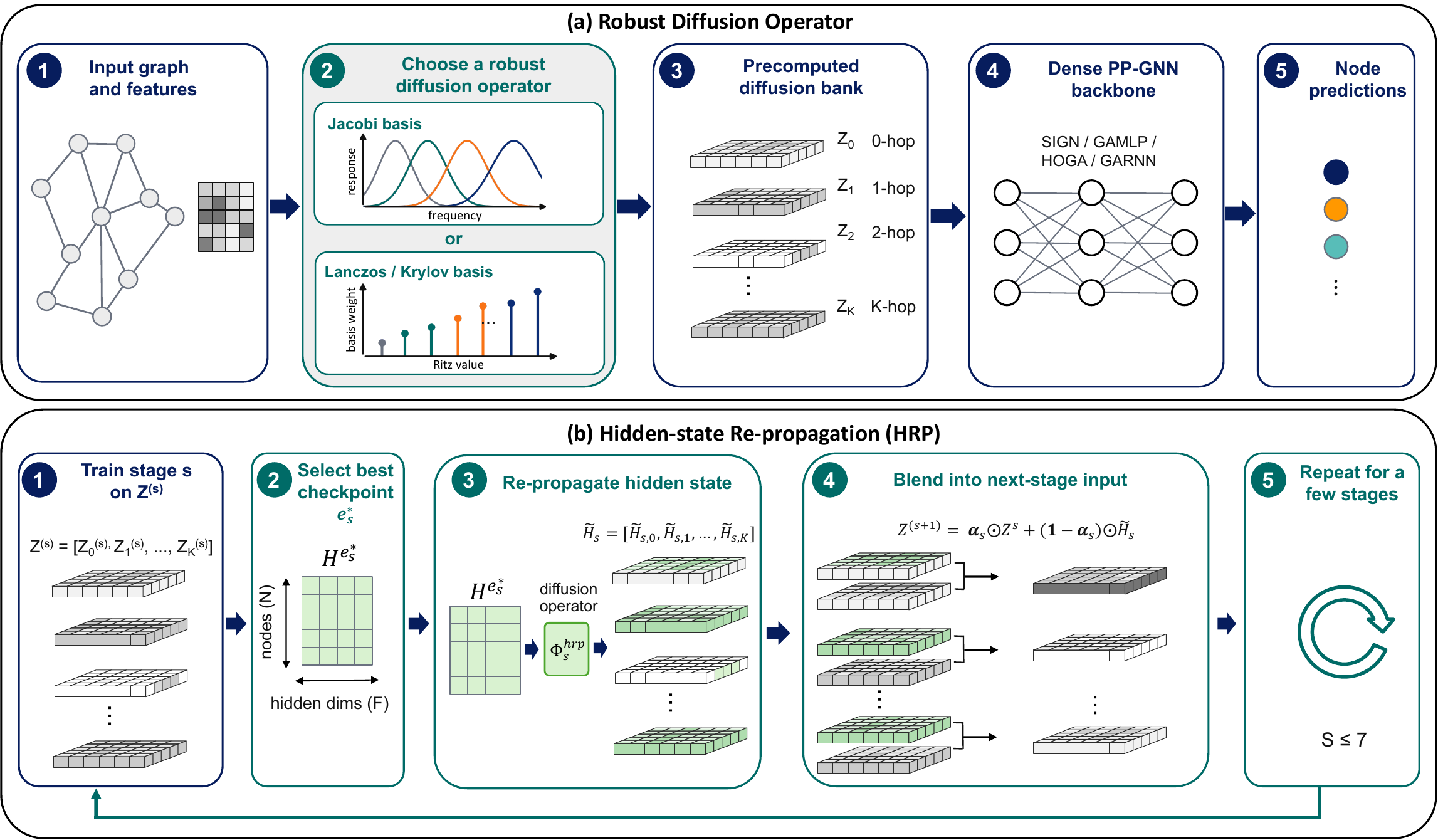}
  \caption{Overview of the proposed robust \PPG framework. 
(a) Robust diffusion operator: the key preprocessing step is to replace the standard monomial hop bank with a better-conditioned diffusion basis, using either a calibrated Jacobi basis or a channel-adaptive Lanczos/Krylov basis. 
This produces a precomputed hop-wise bank $Z=[Z_0,Z_1,\ldots,Z_K]$ that is fed to a dense \PPG backbone for node prediction. 
(b) Hidden-state re-propagation (HRP): at each stage, HRP selects the best hidden representation, re-propagates it to generate a hop-wise hidden-state bank, and blends it with the previous-stage input to form the next-stage bank.
}
  \label{fig:schema}
  \vspace{-5pt}
\end{figure*}
\paragraph{Overview.}
We aim to improve the accuracy of \PPGs while preserving their scalability benefits.
Our approach targets two complementary aspects of standard \PPGs:
(i) the \emph{diffusion operator} used to precompute multi-hop features, and
(ii) the \emph{one-shot} nature of pre-propagation that decouples propagation from evolving representations.
Accordingly, we introduce (1) adoption of stronger graph diffusion operators with more robust diffusion bases, (2) a lightweight \emph{hidden-state re-propagation} mechanism that re-couples propagation with learned representations,
and (3) an optional RNN-based hop aggregator with higher efficiency. The overview of method (1) and (2) are provided in Figure~\ref{fig:schema}.

\subsection{Robust diffusion operators}
\label{sec:method-filters}
\paragraph{Motivation.}
Most \PPGs rely on simple diffusion operators, such as normalized adjacency or random-walk propagation, which typically exhibit low-pass smoothing behavior~\cite{nt2019lowpass, gasteiger2019diffusion}. 
However, heterophily tasks can benefit from preserving or emphasizing higher-frequency and band-pass components~\cite{chien2020gprgnn, bo2021beyond}. 
In principle, combining multi-hop features from a diffusion basis $\{\Phi^k X\}_{k=0}^K$ allows \PPGs to approximate richer spectral responses. 
In practice, however, the standard monomial hop bank is often highly correlated and poorly conditioned~\cite{wang2022powerful}; achieving accurate and stable approximations may therefore require larger hop budgets, which are limited by efficiency and \emph{oversmoothing} considerations~\cite{li2018oversmooth}.

These observations motivate a preprocessing-specific operator-design problem: how to construct a fixed diffusion bank that is well conditioned, spectrally diverse, and reusable throughout dense \PPG training. 
We address this problem by adapting Jacobi-polynomial and Krylov-subspace diffusion bases to the preprocessing regime of \PPGs. 
For Jacobi diffusion, we use an orthogonal polynomial basis on $[-1,1]$ and calibrate its spectral emphasis once from graph-level statistics, reducing the need for validation-driven operator search during preprocessing. 
For Krylov diffusion, we construct feature-channel-adaptive Lanczos bases, which provide orthogonal, signal-aligned diffusion components and remain practical because they are computed once before training rather than recomputed across epochs. 
Together, these \PPG-oriented diffusion bases improve the conditioning and diversity of precomputed hop features, enabling richer spectral responses under a fixed hop budget while preserving the decoupled training pipeline.

\subsubsection{Jacobi operator.}

This diffusion operator constructs hop features using Jacobi polynomials, which form an orthogonal basis on $[-1,1]$ under a weight determined by $(\alpha,\beta)$.
Let $\{P_k^{(\alpha,\beta)}(\cdot)\}_{k\ge 0}$ denote Jacobi polynomials.
We define the $k$-th filtered feature as
\begin{equation}
\label{eq:jacobi_features}
X_k \;=\; P_k^{(\alpha,\beta)}(\tilde L)\,X, \qquad k=0,\dots,K.
\end{equation} where $\tilde L$ is the shifted Laplacian $L - I$ that has spectrum in $[-1,1]$, matching the canonical domain of classical orthogonal polynomials.
Rather than explicitly forming $P_k(\tilde L)$, we compute $X_k$ via the three-term recurrence
\begin{align}
\label{eq:jacobi_recurrence_features}
X_0 &= X, \nonumber\\
X_1 &= \Big(a_0 \tilde L + b_0 I\Big)X_0, \nonumber\\
X_{k+1} &= \Big(a_k \tilde L + b_k I\Big)X_k \;-\; c_k X_{k-1}, \qquad k\ge 1,
\end{align}
where the scalars $a_k,b_k,c_k$ are determined by the Jacobi parameters $(\alpha,\beta)$ (standard closed-form coefficients).
This yields a polynomial diffusion basis that can be better conditioned than the monomial basis $\{\tilde L^k X\}$, and shows better empirical performance \cite{wang2022powerful}.

\textbf{Choice of Jacobi weights.}
The parameters $(\alpha,\beta)$ control how the basis allocates ``resolution'' across the spectral interval $[-1,1]$:
informally, $\alpha$ emphasizes behavior near the high-frequency end ($\lambda\!\approx\!1$) and $\beta$ near the low-frequency end ($\lambda\!\approx\!-1$).
Two common special cases are:
\begin{equation}
\label{eq:special_jacobi}
\begin{aligned}
(\alpha,\beta)=(0,0) &\;\Rightarrow\; \text{Legendre},\\
(\alpha,\beta)=\Big(-\tfrac12,-\tfrac12\Big) &\;\Rightarrow\; \text{Chebyshev}.
\end{aligned}
\end{equation}

Beyond fixed choices, we adapt $(\alpha,\beta)$ to the input graph via a one-time \emph{spectral calibration} step.

\textbf{One-time spectral calibration.}
Although we cannot learn $(\alpha,\beta)$ during training due to the decoupling nature of \PPGs, we can still adapt them to a given graph by performing a lightweight, one-time calibration on $\tilde L$ during preprocessing.
Using stochastic Chebyshev moments, we obtain a coarse estimate of the spectral density on $[-1,1]$. 
We then compare the spectral mass in low- versus high-frequency regions and map their imbalance to Jacobi weights $(\alpha,\beta)$ via a simple monotone rule.
Choosing $(\alpha,\beta)$ this way, rather than through validation-driven search, preserves the \PPG preprocessing contract by avoiding repeated data preparation and keeping the diffusion bank decoupled from training and reusable throughout dense optimization.
Intuitively, the resulting basis allocates more resolution to the frequency region where the graph spectrum concentrates, while keeping calibration fully decoupled from training. Details are provided in Appendix~\ref{app:calibration}.

\subsubsection{Lanczos/Krylov operator.}
\label{subsec:Lanczos}
While the Jacobi operator in \eqref{eq:jacobi_features} uses a \emph{fixed} polynomial basis shared across all feature channels, we additionally consider a \emph{data-adaptive} diffusion basis that (i) is orthogonal by construction and (ii) can capture channel-specific spectral content under a small hop budget.
The key idea is to approximate the spectral action of $\tilde L$ on each channel via a low-rank Krylov subspace projection.

Let $X\in\mathbb{R}^{N\times D}$ be the node feature matrix and let $x_c \in \mathbb{R}^{N}$ denote its $c$-th channel.
For an integer $m \ll N$, define the order-$m$ Krylov subspace with start vector $x_c$
\begin{equation}
\label{eq:krylov_def}
\mathcal{K}_m(\tilde L, x_c)
\;=\;
\mathrm{span}\big\{x_c,\ \tilde L x_c,\ \dots,\ \tilde L^{m-1} x_c\big\}.
\end{equation}
Again, directly using the monomial vectors $\{\tilde L^k x_c\}$ is suboptimal (the vectors become increasingly collinear as $k$ grows).
Instead, assuming $x_c\neq 0$, we run $m$ steps of the Lanczos iteration on the symmetric matrix $\tilde L$
with the normalized start vector $q_{c,1}=x_c/\|x_c\|$ and set $\beta_{c,0}=0$ and $q_{c,0}=0$.
For $j=1,2,\dots,m$, Lanczos constructs scalars $\alpha_{c,j}\in\mathbb{R}$ and $\beta_{c,j}\ge 0$
and the next basis vector $q_{c,j+1}$ via the three-term recurrence
\begin{equation}
\label{eq:lanczos_3term}
\begin{aligned}
r_{c,j} \;&=\; \tilde L q_{c,j} \;-\; \beta_{c,j-1} q_{c,j-1} \;-\; \alpha_{c,j} q_{c,j}, \\
\beta_{c,j} \;&=\; \|r_{c,j}\|,
\qquad
q_{c,j+1} \;=\; r_{c,j}/\beta_{c,j},
\end{aligned}
\end{equation}

where $\alpha_{c,j}=q_{c,j}^\top \tilde L q_{c,j}$.
Stacking $Q_c=[q_{c,1},\dots,q_{c,m}]\in\mathbb{R}^{N\times m}$ yields an orthonormal basis
($Q_c^\top Q_c=I_m$) and a symmetric tridiagonal matrix $T_c\in\mathbb{R}^{m\times m}$
with diagonal entries $\alpha_{c,j}$ and off-diagonal entries
$\beta_{c,j}$, which represents the projection of $\tilde L$ onto
$\mathcal{K}_m(\tilde L,x_c)$.


\smallskip
\noindent\textbf{Low-rank projection.}
Given a spectral response $g(\cdot)$, we approximate the action of $g(\tilde L)$ on $x_c$
by projecting $\tilde L$ onto the Krylov subspace and lifting the result back:
\begin{equation}
\label{eq:proj_spectral_action_general}
g(\tilde L)\,x_c
\;\approx\;
Q_c\, g(T_c)\, Q_c^\top x_c,
\qquad
T_c \;=\; Q_c^\top \tilde L Q_c,
\end{equation}
Consider the eigendecomposition of $T_c$:
\begin{equation}
\label{eq:T_eig}
T_c \;=\; U_c \Lambda_c U_c^\top,
\qquad
U_c=[u_{c,1},\dots,u_{c,m}]\in\mathbb{R}^{m\times m},
\end{equation}
where $\Lambda_c=\mathrm{diag}(\lambda_{c,1},\dots,\lambda_{c,m})$ contains the Ritz values.
Then we choose $g$ to \emph{select a single eigenmode} of $T_c$, which yields an explicit decomposition into decorrelated components. Specifically, for the $i$-th mode define a response that keeps only the Ritz value $\lambda_{c,i}$:
\begin{equation}
\label{eq:mode_selector}
g_i(\Lambda_c)=\mathrm{diag}(0,\dots,0,1,0,\dots,0),
\qquad
g_i(T_c)=u_{c,i}u_{c,i}^\top
\end{equation}
Combining with \eqref{eq:proj_spectral_action_general} gives the $i$-th \emph{Ritz component} of $x_c$:
\begin{equation}
\label{eq:ritz_component_proj}
z_{c,i}
\;:=\;
Q_c\, g_i(T_c)\, Q_c^\top x_c
\;=\;
Q_c (u_{c,i}u_{c,i}^\top) Q_c^\top x_c
\end{equation}
Because the Lanczos start vector is $q_{c,1}=x_c/\|x_c\|$, we have $Q_c^\top x_c=\|x_c\|e_1$, and thus
\begin{equation}
\label{eq:ritz_component_scalar}
z_{c,i}
\;=\;
\|x_c\|\,(u_{c,i}^\top e_1)\, Q_c u_{c,i}
\;=\;
(\|x_c\|\cdot u_{c,i,1})\, y_{c,i},
\end{equation}
where $u_{c,i,1}$ is the first entry of $u_{c,i}$, and $y_{c,i}$ is the \emph{Ritz vectors} in the original node space. 

\paragraph{Interpretation.}
The set $\{z_{c,i}\}_{i=1}^m$ forms a channel-wise \emph{Ritz-component bank}: it decomposes $x_c$ into $m$ diffusion components within $\mathcal{K}_m(\tilde L,x_c)$, with mode shapes given by the lifted Ritz vectors $\{y_{c,i}\}$ and associated ``frequencies'' given by the Ritz values in $\Lambda_c$. 
Compared with the standard monomial hop bank $\{x_c,\tilde Lx_c,\ldots,\tilde L^{m-1}x_c\}$, this component bank provides a better coordinate system for \PPG aggregation. 
Its orthogonality reduces correlations among hop features, its channel-specific start vector aligns the basis with the signal being filtered, and each response coefficient $g(\lambda_{c,i})$ directly weights one spectral component. 
These properties make the resulting diffusion bank easier for \PPG heads to aggregate under a fixed hop budget.

\smallskip
\noindent\textbf{Cost and practicality.}
When Lanczos is run per channel, each iteration applies the same sparse operator $\tilde L$ to the current
vectors; in practice, we batch all channels and compute $\tilde L Q_j$ using a single SpMM per step.
Each step therefore costs $O(\mathrm{nnz}(\tilde L)\,D)$, plus $O(ND)$ columnwise dot products and
scalings to form $\alpha_{c,j}$, $\beta_{c,j}$, and normalize $q_{c,j+1}$.
On sparse graphs with average degree $\bar d=\mathrm{nnz}/N$, the $O(ND)$ term is typically dominated by the
SpMM, yielding an overall complexity $O(m\,\mathrm{nnz}(\tilde L)\,D)$, the same leading order as $m$ steps of
standard diffusion preprocessing.
We also eigendecompose the small tridiagonal matrix $T_c\in\mathbb{R}^{m\times m}$ per channel,
which costs $O(D\,m^3)$ and is negligible for $m\le 15$ used in our settings.
Forming all Ritz vectors via $Y_c = Q_c U_c$ costs $O(N m^2)$ per channel, i.e., $O(N m^2 D)$ overall, comparable to the $m$ steps of diffusion cost.


\subsection{Few-shot hidden-state re-propagation}
\label{sec:method-reprop}

\paragraph{Motivation.}
Even with stronger diffusion operators, the dense backbones of \PPGs can only learn on a fixed set of precomputed hop features.
In contrast, \MPGs repeatedly interleave propagation with nonlinear transformations, so \emph{task-adapted} representations are iteratively diffused and refined.
To tackle this limitation, we introduce a lightweight \emph{few-shot} re-propagation mechanism: we occasionally re-apply diffusion to intermediate hidden states during training, and feed the diffused representations back to the backbone, bridging one-shot \PPGs and iterative propagation while preserving dense-dominated training.

\paragraph{Re-propagation mechanism.}

Let $Z^{(s)}$ denote the hop-wise backbone input at stage $s$, and let $H^{(e)}\in\mathbb{R}^{N\times d}$ be the hidden states \emph{before} the final output projection at epoch $e$.
We split training into $S$ stages; stage $s$ runs for $E_s$ epochs and is long enough for validation performance to stabilize.
At the end of stage $s$, we select an epoch index $e_s^\star$ 
(default: the epoch attaining the best validation accuracy within the stage), 
detach the corresponding hidden representation 
$\bar H_s=\operatorname{stopgrad}(H^{(e_s^\star)})$, 
and re-propagate it to form a hop-wise hidden-state bank:
\begin{equation}
\widetilde H_s
=
[\widetilde H_{s,0},\widetilde H_{s,1},\ldots,\widetilde H_{s,K}]
=
\mathcal{B}_{K}(\bar H_s;\Phi_s^{\mathrm{hrp}}),
\label{eq:method-reprop-core}
\end{equation}
where $\Phi_s^{\mathrm{hrp}}$ denotes the HRP diffusion operator and 
$\mathcal{B}_{K}(\cdot;\Phi_s^{\mathrm{hrp}})$ denotes the corresponding $K$-hop diffusion-bank construction.
For example, $\mathcal{B}_{K}$ can be instantiated either by a power-form bank 
$[\bar H_s,\Phi_s^{\mathrm{hrp}}\bar H_s,\ldots,(\Phi_s^{\mathrm{hrp}})^K\bar H_s]$ 
or by a Lanczos/Krylov-derived bank as in Section~\ref{subsec:Lanczos}.
We then augment the backbone input for the \emph{next} stage by blending the previous-stage hop-wise input with the re-propagated hidden-state bank:
\begin{equation}
\begin{aligned}
Z^{(s+1)} 
&=\; \boldsymbol{\alpha}_s \odot Z^{(s)}
\;+\; 
\bigl(\mathbf{1}-\boldsymbol{\alpha}_s\bigr)\odot \widetilde{H}_s,\\
H^{(e)}
&=\; f_\theta\!\left(Z^{(s)}\right),
\qquad \text{for } e \in \text{stage } s.
\end{aligned}
\label{eq:method-reprop-aug}
\end{equation}

with $Z^{(1)}\!=\!Z$. 
Here $\boldsymbol{\alpha}_s\in[0,1]^{K+1}$ is a hop-wise vector, and we set $\alpha_{s,0}=1$ so that the $0$-hop feature is kept unchanged, and blend every propagated hop $k\geq 1$ with its re-propagated hidden-state counterpart. 
The propagated-hop weights can be shared or hop-specific, and we optionally apply a cosine decay across stages so later stages place less weight on the re-propagated bank.

\paragraph{Optional auto search.}
We optionally enable a lightweight auto-search to reduce manual tuning of HRP.
Specifically, we (i) tune the stage-wise blending weights with a simple evolution strategy, and (ii) select a stage checkpoint and re-propagation operator from a small candidate pool using a tight screening budget.
We further promote diverse candidates by scoring how \emph{spectrally different} a hidden state is from the original input via per-channel spectral moments of the normalized Laplacian.
Additional details are provided in Appendix~\ref{app:auto-search}.


\paragraph{Efficiency.}
Re-propagation occurs at low frequency ($S\le 7$ in all experiments).
Thus training remains dominated by dense computation in $f_\theta$, while diffusion is amortized across stages, preserving the scalability advantages of \PPGs.



\subsection{RNN-based hop aggregator}
\label{sec:method-rnn}

\paragraph{Motivation.}
A critical design choice in \PPGs is how to aggregate hop-wise features.
Existing backbones differ substantially: SIGN \cite{frasca2020sign} concatenates $\{Z_k\}_{k=0}^{K}$ and applies an MLP; GAMLP \cite{zhang2022gamlp} performs stage-wise attention over hops to emphasize informative diffusions; and HOGA \cite{deng2024hoga} models the hop axis as a sequence and applies multi-head attention (MHA).
When $K$ is moderately large, MHA-like aggregators can incur non-trivial overhead.
Motivated by the empirical observation that recurrent models can match MHA accuracy on short sequences at lower costs, we introduce an efficient RNN-based hop aggregator that treats hops as a sequence, following the sequential view in HOGA but with reduced compute and memory.

\paragraph{Architecture.}
Let $Z_k\in\mathbb{R}^{N\times d}$ denote the $k$-hop feature; we apply a shared recurrent update along the hop dimension:
\begin{equation}
S_k \;=\; \operatorname{RNN}_\theta(Z_k, S_{k-1}),\qquad k=0,\ldots,K,
\label{eq:method-rnn-update}
\end{equation}
where $S_k\in\mathbb{R}^{N\times h}$ is the recurrent state.
We form the aggregated representation either from the last state or from a lightweight pooling over states:
\begin{equation}
Z_{\text{agg}} \;=\; S_K
\quad\text{or}\quad
Z_{\text{agg}} \;=\; \operatorname{Pool}\!\left(\{S_k\}_{k=0}^{K}\right),
\label{eq:method-rnn-readout}
\end{equation}
and feed $Z_{\text{agg}}$ to the dense predictor (MLP) for supervision.
This hop aggregator is \emph{optional} and can be paired with any diffusion family and with hidden-state re-propagation.

\section{Experimental Setup}
We conduct extensive experiments on both heterophilic and homophilic node classification benchmarks.
We compare against a wide collection of \MPG baselines and show that, after incorporating our proposed diffusion operator suite and few-shot hidden-state re-propagation (HRP) across multiple training stages, \PPGs can match or surpass \MPGs while retaining the scalability benefits.

\subsection{Datasets}
We evaluate on six heterophilic graphs from \cite{lim2021hetero, platonov2023critical}, which were curated to address common issues in earlier heterophilic benchmarks (e.g., unstable evaluation and split sensitivity).
To verify that our approach is not restricted to heterophily, we additionally report results on six widely used homophilic datasets \cite{shchur1811pitfalls, hu2020ogb}.
Dataset statistics and detailed split settings are provided in Appendix~\ref{app:dataset}.

\subsection{Baselines}
\paragraph{Message-passing GNNs.}
We compare against a broad set of standard message-passing GNN baselines (e.g., GCN/GAT/GraphSAGE and heterophily-oriented variants). For readability, we defer the mapping from baseline names to their original papers to Appendix~\ref{app:gnn-baselines}, as well as their hyperparameter settings. 

\paragraph{Pre-propagation GNNs.}
We evaluate on three representative \PPG baselines—SIGN~\cite{frasca2020sign}, HOGA~\cite{deng2024hoga}, and GAMLP~\cite{zhang2022gamlp}—together with our RNN-based hop aggregator, \emph{graph-augmented RNN} (GARNN).
These models span different dense backbones, including MLP-style hop aggregation and attention-based hop aggregation.
Our framework applies to each backbone by (i) enabling few-shot hidden-state re-propagation (HRP) across training stages and (ii) replacing the diffusion operator suite used for both feature preprocessing and HRP.
For GAMLP, we use GAMLP-R without label inputs by default, unless stated otherwise. Detailed settings are provided in Appendix~\ref{app:ppgnnbaseline} and \ref{app:operators}.


\section{Evaluation}
\label{sec:eval}
We evaluate our two contributions---\emph{robust diffusion operators} and \emph{few-shot hidden-state re-propagation (HRP)}---by integrating them into representative pre-propagation \PPG baselines and comparing against competitive \MPG models.
Our evaluation is organized around four questions:
(i) do the proposed diffusion families improve \PPGs under the same hop budget,
(ii) does re-propagation of intermediate representations further narrow the gap to iterative message passing,
(iii) how does HRP compare to label-based alternatives such as using labels as input or label propagation, and
(iv) what accuracy--efficiency trade-offs do these components introduce?
We additionally provide ablations to isolate the effect of each module.

\subsection{Results on heterophilic graphs}
\begin{table}[t]
  \centering
  \captionsetup{width=\columnwidth}
  \caption{Average node classification results over $10$ runs on heterophilic datasets ($+$: apply robust diffusion operators and HRP to \PPG baselines). Accuracy is reported for \texttt{roman-empire} and \texttt{amazon-ratings}, and ROC AUC is reported for \texttt{minesweeper}, \texttt{tolokers}, and \texttt{questions}.}
  \vspace{-6pt}
  \label{tab:hetero}
  \setlength{\tabcolsep}{3.2pt} 
  \renewcommand{\arraystretch}{0.95} 
  \scriptsize
  \resizebox{\columnwidth}{!}{%
  \begin{tabular}{lccccc}
  \toprule
   & \makecell[c]{roman\\-empire} & \makecell[c]{amazon\\-ratings} & \makecell[c]{mine\\sweeper} & \makecell[c]{tolo\\-kers} & \makecell[c]{ques\\-tions} \\
  \midrule
  GCN          & $73.7 \pm 0.7$ & $48.7 \pm 0.6$ & $89.8 \pm 0.5$ & $83.6 \pm 0.7$ & $76.1 \pm 1.3$ \\
  SAGE    & $85.7 \pm 0.7$ & $53.6 \pm 0.4$ & $93.5 \pm 0.6$ & $82.4 \pm 0.4$ & $76.4 \pm 0.6$ \\
  GAT-sep      & $88.8 \pm 0.4$ & $52.7 \pm 0.6$ & $\mathbf{93.9 \pm 0.4}$ & $83.8 \pm 0.4$ & $76.8 \pm 0.7$ \\
  H2GCN        & $60.1 \pm 0.5$ & $36.5 \pm 0.2$ & $89.7 \pm 0.3$ & $73.4 \pm 1.0$ & $63.6 \pm 1.5$ \\
  GPRGNN       & $64.8 \pm 0.3$ & $44.9 \pm 0.3$ & $86.2 \pm 0.6$ & $72.9 \pm 1.0$ & $55.5 \pm 0.9$ \\
  FSGNN        & $79.9 \pm 0.6$ & $52.7 \pm 0.8$ & $90.1 \pm 0.7$ & $82.8 \pm 0.6$ & $78.9 \pm 0.9$ \\
  ACMGNN        & $72.7 \pm 0.8$ & $52.7 \pm 0.3$ & $90.7 \pm 0.6$ & $82.1 \pm 0.6$ & $77.4 \pm 1.5$ \\
  GloGNN       & $59.6 \pm 0.7$ & $36.9 \pm 0.1$ & $51.1 \pm 1.2$ & $73.4 \pm 1.2$ & $65.7 \pm 1.2$ \\
  GGCN         & $74.5 \pm 0.5$ & $43.0 \pm 0.3$ & $87.5 \pm 1.2$ & $77.3 \pm 1.1$ & $71.1 \pm 1.6$ \\
  O-GNN   & $77.7 \pm 0.4$ & $47.3 \pm 0.7$ & $80.6 \pm 1.1$ & $75.6 \pm 1.4$ & $75.1 \pm 1.0$ \\
  G$^2$-GNN    & $82.2 \pm 0.8$ & $47.9 \pm 0.6$ & $91.8 \pm 0.6$ & $82.5 \pm 0.8$ & $74.8 \pm 0.9$ \\
  DIR-GNN      & $\mathbf{91.2 \pm 0.3}$ & $47.9 \pm 0.4$ & $87.0 \pm 0.7$ & $81.2 \pm 1.1$ & $76.1 \pm 1.2$ \\
  tGNN         & $80.0 \pm 0.8$ & $48.2 \pm 0.5$ & $91.9 \pm 0.8$ & $70.8 \pm 1.8$ & $76.4 \pm 1.8$ \\
  \midrule
  SIGN         & $80.0 \pm 0.5$ & $54.1 \pm 0.7$ & $90.7 \pm 0.6$ & $84.1 \pm 1.0$ & $78.6 \pm 1.1$ \\
  HOGA         & $79.4 \pm 0.6$ & $51.6 \pm 0.3$ & $90.5 \pm 0.7$ & $78.1 \pm 0.8$ & $78.3 \pm 1.0$ \\
  GAMLP        & $78.9 \pm 0.7$ & $52.2 \pm 0.4$ & $90.5 \pm 0.7$ & $85.1 \pm 0.8$ & $75.9 \pm 1.3$ \\
  GARNN        & $79.2 \pm 0.7$ & $53.2 \pm 0.6$ & $90.3 \pm 0.7$ & $82.1 \pm 0.8$ & $78.4 \pm 1.1$ \\
  \midrule
  \rowcolor{gray!15} SIGN$^{+}$         & $81.4 \pm 1.1$ & $\mathbf{54.4 \pm 0.9}$ & $91.3 \pm 0.7$ & $84.6 \pm 1.0$ & $78.4 \pm 1.1$ \\
  \rowcolor{gray!15} HOGA$^{+}$         & $85.5 \pm 1.3$ & $53.8 \pm 0.7$ & $92.1 \pm 0.7$ & $83.6 \pm 0.6$ & $\mathbf{79.0 \pm 0.8}$ \\
  \rowcolor{gray!15} GAMLP$^{+}$        & $83.1 \pm 0.6$ & $52.2 \pm 0.4$ & $92.1 \pm 0.8$ & $\mathbf{85.5 \pm 0.5}$ & $78.2 \pm 1.4$ \\
  \rowcolor{gray!15} GARNN$^{+}$        & $85.5 \pm 0.5$ & $53.6 \pm 0.7$ & $92.3 \pm 0.5$ & $84.0 \pm 0.6$ & $78.5 \pm 1.2$ \\
  \bottomrule
  \end{tabular}
  }
  \vspace{-6pt}
  \end{table}
  
\begin{table}[ht]
  \centering
  \caption{Average node classification accuracy (\%) $\pm$ std over $5$ runs on large datasets ($+$: apply robust diffusion operators and HRP to \PPG baselines).}
  \vspace{-5pt}
  \label{tab:large2}
  \small
  \scalebox{0.9}{
  \begin{tabular}[t]{lcc}
  \toprule
     &  ogbn-arxiv   &  pokec \\
  \midrule
  GCN  & $71.74 \pm 0.29$  & $75.45 \pm 0.17$  \\
  SAGE & $71.49 \pm 0.27$  & $78.40 \pm0.45$  \\
  GAT   &  \textbf{$72.01 \pm 0.20$}  & $81.52 \pm 0.17$  \\
  GPRGNN   &  $71.10 \pm 0.12$  & $78.83 \pm 0.05$  \\
  LINKX  &  $66.18 \pm 0.33$  & \textbf{$82.04\pm 0.07$}  \\
  \midrule
  SIGN   &  $71.79 \pm 0.14$  & $81.02 \pm 0.33$  \\
  HOGA   &  $72.21 \pm 0.35$  & $82.11 \pm 0.25$  \\
  GAMLP  &  $71.09 \pm 0.36$  & $ 78.25\pm 0.69$  \\
  GARNN  &  $72.04 \pm 0.12$  & $80.90 \pm 0.43$  \\
  \midrule
  \rowcolor{gray!15} SIGN$^{+}$   &  $72.56 \pm 0.24$  & $83.85 \pm 0.46$  \\
  \rowcolor{gray!15} HOGA$^{+}$   &  $72.35 \pm 0.16$  & $84.50 \pm 0.12$  \\
  \rowcolor{gray!15} GAMLP$^{+}$  &  $\mathbf{72.62 \pm 0.18}$  & $83.68 \pm 0.58$  \\
  \rowcolor{gray!15} GARNN$^{+}$  &  $72.05 \pm 0.37$  & $\mathbf{84.63 \pm 0.08}$  \\
  \bottomrule
  \end{tabular}
  }
  \end{table}
Table~\ref{tab:hetero} and Table~\ref{tab:large2} report node classification performance on six commonly used heterophilic graphs,
including five small graphs and the large \texttt{pokec} graph.
For each \PPG baseline, results \emph{without} the ``$+$'' annotation use standard diffusion operators
(either normalized adjacency or random-walk diffusion), while results \emph{with} ``$+$'' apply our full method:
robust diffusion operators together with HRP.
Before applying our techniques, \PPGs exhibit a clear accuracy gap to \MPGs on challenging datasets such as
\texttt{\seqsplit{roman-empire}} and \texttt{\seqsplit{minesweeper}}, reaching up to $11\%$ absolute test accuracy.
Our method consistently improves \PPGs and reduces this gap substantially---often by roughly half---bringing \PPGs close to strong \MPG baselines such as GraphSAGE and GAT.
On the remaining heterophilic datasets, our method provides additional gains on top of already competitive \PPGs, and
achieves higher test accuracy than \MPGs on $4$ out of $6$ heterophilic graphs.
Overall, across the $6$ heterophilic datasets, our full method improves test performance over the corresponding \PPG baselines by $\mathbf{+2.18}$ points on average (max $\mathbf{+6.3}$).

\subsection{Results on homophilic graphs}
\begin{table}[t]
  \centering
  \captionsetup{width=\columnwidth}
  \caption{Average node classification accuracy (\%) $\pm$ std over $10$ runs on homophilic datasets ($+$: apply robust diffusion operators and HRP to \PPG baselines).}
  \vspace{-6pt}
  \label{tab:homo}
  \setlength{\tabcolsep}{3.2pt}
  \renewcommand{\arraystretch}{0.95}
  \scriptsize
  \resizebox{\columnwidth}{!}{%
  \begin{tabular}{lccccc}
  \toprule
   & \makecell[c]{Computer} & \makecell[c]{Photo} & \makecell[c]{CS} & \makecell[c]{Physics} & \makecell[c]{WikiCS} \\
  \midrule
 GCN          & $89.7 \pm 0.5$ & $92.7 \pm 0.2$ & $92.9 \pm 0.1$ & $96.2 \pm 0.1$ & $77.5 \pm 0.9$ \\
 SAGE    & $91.2 \pm 0.3$ & $94.6 \pm 0.1$ & $93.9 \pm 0.1$ & $96.5 \pm 0.1$ & $74.8 \pm 1.0$ \\
 GAT          & $90.8 \pm 0.1$ & $93.9 \pm 0.1$ & $93.6 \pm 0.1$ & $96.2 \pm 0.1$ & $76.9 \pm 0.8$ \\
 GCNII        & $91.0 \pm 0.4$ & $94.3 \pm 0.2$ & $92.2 \pm 0.1$ & $96.0 \pm 0.1$ & $78.7 \pm 0.6$ \\
 GPRGNN       & $89.3 \pm 0.3$ & $94.5 \pm 0.1$ & $95.1 \pm 0.1$ & $96.9 \pm 0.1$ & $78.1 \pm 0.2$ \\
 APPNP        & $90.2 \pm 0.2$ & $94.3 \pm 0.1$ & $94.5 \pm 0.1$ & $96.5 \pm 0.1$ & $78.9 \pm 0.1$ \\
 PPRGo        & $88.7 \pm 0.2$ & $93.6 \pm 0.1$ & $92.5 \pm 0.2$ & $95.5 \pm 0.1$ & $77.9 \pm 0.4$ \\
 GGCN         & $91.8 \pm 0.2$ & $94.5 \pm 0.1$ & $\mathbf{95.3 \pm 0.1}$ & $\mathbf{97.1 \pm 0.1}$ & $78.4 \pm 0.5$ \\
 O-GNN   & $92.0 \pm 0.1$ & $95.1 \pm 0.2$ & $95.0 \pm 0.1$ & $97.0 \pm 0.1$ & $79.0 \pm 0.7$ \\
 tGNN         & $83.4 \pm 1.3$ & $89.9 \pm 0.7$ & $92.9 \pm 0.5$ & $96.2 \pm 0.2$ & $71.5 \pm 1.1$ \\
  \midrule
  SIGN         & $84.2 \pm 0.3$ & $90.5 \pm 0.7$ & $93.9 \pm 0.4$ & $95.5 \pm 0.2$ & $79.2 \pm 0.7$ \\
  HOGA         & $84.5 \pm 0.8$ & $91.9 \pm 0.7$ & $94.6 \pm 0.3$ & $95.9 \pm 0.3$ & $78.9 \pm 0.5$ \\
  GAMLP        & $86.8 \pm 0.3$ & $91.2 \pm 0.9$ & $94.8 \pm 0.3$ & $96.3 \pm 0.2$ & $\mathbf{80.7 \pm 0.7}$ \\
  GARNN        & $85.3 \pm 0.9$ & $92.1 \pm 0.7$ & $94.5 \pm 0.4$ & $96.0 \pm 0.2$ & $78.9 \pm 0.7$ \\
  \midrule
  \rowcolor{gray!15} SIGN$^{+}$         & $91.7 \pm 0.5$ & $94.3 \pm 0.7$ & $94.7 \pm 0.3$ & $96.1 \pm 0.2$ & $79.4 \pm 0.7$ \\
  \rowcolor{gray!15} HOGA$^{+}$         & $91.4 \pm 0.7$ & $94.2 \pm 0.6$ & $95.1 \pm 0.3$ & $96.6 \pm 0.2$ & $79.5 \pm 0.3$ \\
  \rowcolor{gray!15} GAMLP$^{+}$        & $91.6 \pm 0.6$ & $\mathbf{95.3 \pm 0.6}$ & $94.9 \pm 0.4$ & $96.5 \pm 0.3$ & $80.6 \pm 0.4$ \\
  \rowcolor{gray!15} GARNN$^{+}$        & $\mathbf{92.1 \pm 0.3}$ & $94.8 \pm 0.7$ & $94.9 \pm 0.4$ & $96.9 \pm 0.2$ & $79.7 \pm 0.5$ \\
  \bottomrule
  \end{tabular}%
  }
  \vspace{-10pt}
  \end{table}
  
Although motivated by heterophily, our techniques also improve performance on homophilic graphs.
Table~\ref{tab:homo} reports results on five small homophilic datasets, and Table~\ref{tab:large2} additionally includes \texttt{ogbn-arxiv}.
Compared to heterophilic graphs, the baseline accuracy gap between \PPGs and \MPGs is smaller,
which is consistent with the fact that commonly used diffusion operators (normalized adjacency / random walk) behave largely as low-pass smoothers and thus better match homophily.
Nevertheless, we observe substantial gains from our method, with up to $6.8\%$ absolute improvement on \texttt{\seqsplit{amazon-computer}},
and our enhanced \PPGs outperform \MPGs on $3$ out of $6$ homophilic datasets.
Overall, across the $6$ homophilic datasets, our full method improves test accuracy over the corresponding \PPG baselines by $\mathbf{+1.96}$ points on average (max $\mathbf{+7.5}$).

\subsection{Ablating robust operators and HRP}
\begin{table}[t]
  \centering
  \captionsetup{width=\columnwidth}
  \caption{Ablation study (val/test accuracy in \%) for different components in SIGN, HOGA, GAMLP, and GARNN on \texttt{pokec}.}
  \vspace{-6pt}
  \label{tab:ablation}
  \setlength{\tabcolsep}{5.5pt}
  \renewcommand{\arraystretch}{0.95}
  \scriptsize
  \begin{tabular}{lcccc}
  \toprule
  Method & Val Acc. (\%) & Val $\uparrow$ (\%) & Test Acc. (\%) & Test $\uparrow$ (\%) \\
  \midrule
  SIGN & $81.01 \pm 0.32$ & $+0.00$ & $81.02 \pm 0.33$ & $+0.00$ \\
  \quad + robust op. & $83.11 \pm 0.05$ & $+2.10$ & $83.10 \pm 0.05$ & $+2.08$ \\
  \quad + HRP & $83.93 \pm 0.48$ & $+0.82$ & $83.85 \pm 0.46$ & $+0.75$ \\
  \midrule
  HOGA & $82.11 \pm 0.27$ & $+0.00$ & $82.11 \pm 0.25$ & $+0.00$ \\
  \quad + robust op. & $82.68 \pm 0.13$ & $0.57$ & $82.65 \pm 0.10$ & $0.54$ \\
  \quad + HRP & $84.54 \pm 0.11$ & $1.86$ & $84.50 \pm 0.12$ & $1.85$ \\
  \midrule
  GAMLP & $78.26 \pm 0.75$ & $+0.00$ & $78.25 \pm 0.69$ & $+0.00$ \\
  \quad + robust op. & $81.72 \pm 0.42$ & $+3.46$ & $81.28 \pm 0.51$ & $+3.03$ \\
  \quad + HRP & $83.69 \pm 0.58$ & $+1.97$ & $83.68 \pm 0.58$ & $+2.4$ \\
  \midrule
  GARNN & $80.92 \pm 0.38$ & $+0.00$ & $80.90 \pm 0.43$ & $+0.00$ \\
  \quad + robust op. & $83.15 \pm 0.05$ & $+2.23$ & $83.19 \pm 0.05$ & $+2.29$ \\
  \quad + HRP & $84.66 \pm 0.06$ & $+1.51$ & $84.63 \pm 0.08$ & $+1.44$ \\
  \bottomrule
  \end{tabular}
  \vspace{-10pt}
\end{table}
We conduct ablations on the large \texttt{pokec} dataset to quantify the contribution of each component (Table~\ref{tab:ablation}).
Rows labeled ``$+$ robust op.'' replace the standard diffusion operator with our proposed operator families, while rows labeled ``HRP'' further enable hidden-state re-propagation.
Across four \PPG backbones, the robust diffusion operators improve validation/test accuracy by \textbf{2.09}\%/\textbf{1.99}\% on average,
and HRP provides an additional \textbf{1.54}\%/\textbf{1.61}\% gain.
These results indicate that (i) richer diffusion bases are beneficial even without iterative refinement, and (ii) re-propagating intermediate representations provides complementary improvements beyond stronger preprocessing alone.

\subsection{Comparing HRP with label propagation}
\begin{table}[t]
  \centering
  \captionsetup{width=\columnwidth}
  \caption{Label strategy comparison (val/test accuracy in \%) for GAMLP and GARNN on \texttt{roman-empire}.}
  \vspace{-6pt}
  \label{tab:label_cmp}
  \setlength{\tabcolsep}{5.5pt}
  \renewcommand{\arraystretch}{0.95}
  \scriptsize
  \begin{tabular}{lcc}
  \toprule
  Method & Val Acc. (\%) & Test Acc. (\%) \\
  \midrule
  \multicolumn{3}{l}{\textbf{GAMLP}} \\
  no label & $81.72 \pm 0.42$ & $81.28 \pm 0.51$ \\
  \quad - label-as-input & $81.29 \pm 0.71$ & $80.75 \pm 0.66$ \\
  \quad - label prop. & $81.98 \pm 0.31$ & $81.53 \pm 0.56$ \\
  \quad - HRP & $83.53 \pm 0.63$ & $83.08 \pm 0.60$ \\
 
  \midrule
  \multicolumn{3}{l}{\textbf{GARNN}} \\
  no label & $82.50 \pm 0.44$ & $82.06 \pm 0.42$ \\
  \quad - label-as-input & $82.72 \pm 0.36$ & $82.28 \pm 0.73$ \\
  \quad - label prop. & $83.23 \pm 0.37$ & $82.62 \pm 0.40$ \\
  \quad - HRP & $86.11 \pm 0.55$ & $85.53 \pm 0.48$ \\
  
  \bottomrule
  \end{tabular}
  \vspace{-10pt}
\end{table}
A natural question is whether hidden-state re-propagation (HRP) simply replicates label-based techniques such as using labels as input or label propagation.
We therefore compare HRP against two label strategies on \texttt{roman-empire}:
(i) \emph{label-as-input}, where we encode training labels as one-hot vectors and mask validation/test labels following standard practice, and
(ii) \emph{label propagation / self-training}, following the label propagation procedure used in GAMLP, where we augment inputs with soft predictions (logits) from the previous stage under knowledge distillation guidance.

As shown in Table~\ref{tab:label_cmp}, label-as-input and label propagation yield only marginal improvements over the no-label baseline, whereas HRP provides a substantially larger gain for both GAMLP and GARNN.
Moreover, on heterophilic graphs, labels are typically less smooth with respect to the graph structure, making propagation-based methods harder to leverage and sometimes even detrimental (e.g., the ``label-as-input'' variant in Table~\ref{tab:label_cmp}).
This motivates HRP: instead of propagating potentially noisy label signals, HRP re-propagates intermediate hidden representations, which (i) are higher-dimensional and can encode richer, task-relevant relational information beyond class probabilities, and (ii) avoid introducing a direct label feature channel that can act as a shortcut to the target.

\subsection{Training efficiency impact}
\begin{table}[t]
  \centering
  \captionsetup{width=\columnwidth}
  \caption{Wall-clock end-to-end training time breakdown for HOGA and GARNN on \texttt{pokec}. Stage 1 represents runtime without HRP.}
  \label{tab:training_efficiency}
  \setlength{\tabcolsep}{6pt}
  \renewcommand{\arraystretch}{0.95}
  \scriptsize
  \begin{tabular}{lcccc}
  \toprule
  Method & Stage 1 (sec) & HRP (sec) & Total (sec) & Epoch time (sec) \\
  \midrule
  GARNN & 1272 & 51 & 2290 & 2.4\\
  HOGA & 1240 & 94 & 4293 & 8.4\\
  \bottomrule
  \end{tabular}
  \vspace{-10pt}
\end{table}
\begin{figure}[ht]
    \centering
    \includegraphics[width=0.9\columnwidth]{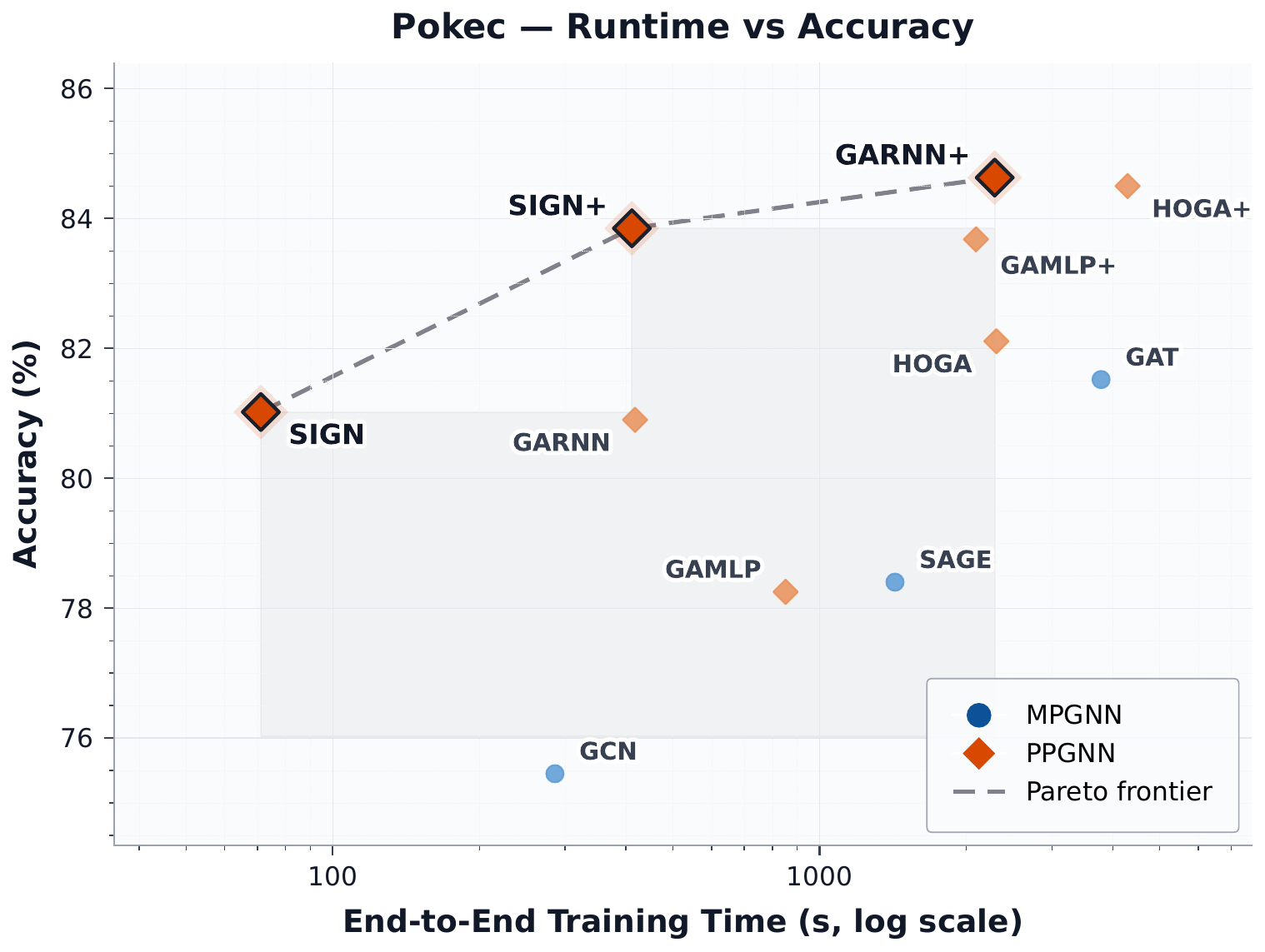}
    \vspace{-5pt}
    \caption{Accuracy--runtime trade-off on \texttt{pokec}. The x-axis reports end-to-end training time in seconds on a logarithmic scale, and the y-axis reports test accuracy. Enhanced \PPG variants with robust diffusion operators and HRP move the \PPG family toward a stronger Pareto frontier, achieving substantially higher accuracy than vanilla \PPGs and offering a favorable accuracy--time trade-off relative to representative \MPG baselines.}
    \label{fig:pareto}
    \vspace{-5pt}
\end{figure}
We next evaluate the training-efficiency impact of the proposed components. 
We next evaluate the efficiency impact of the proposed methods. Jacobi diffusion has similar preprocessing cost to standard normalized-adjacency/random-walk diffusion, while the Lanczos operator adds a Ritz-mode step but remains comparable to a small number of diffusion steps (Section~\ref{sec:method-filters}). HRP introduces a few extra stages, each requiring only a small number of diffusion operations on hidden states; we observe at most $7$ stages overall (typically $<4$). On \texttt{pokec}, Table~\ref{tab:training_efficiency} reports wall-clock end-to-end training time with $4$ stages: total time is about $2\times$ the baseline (stage~1), mainly attributed to extra stages introduced. HRP diffusion accounts for only about $2\%$ of the total training time. It also shows that GARNN trains faster than HOGA while achieving similar accuracy, indicating that RNN is a more efficient hop aggregation choice.

Figure~\ref{fig:pareto} further summarizes the accuracy--runtime trade-off on \texttt{pokec}. 
The proposed components, including robust diffusion operators and HRP, shift the \PPG family toward stronger Pareto-frontier positions: they substantially improve accuracy over vanilla \PPGs while preserving competitive end-to-end training-time trade-offs relative to representative \MPG baselines.

\subsection{Summary}
Overall, robust diffusion operators consistently improve \PPG performance, and HRP further narrows the gap to iterative message passing.
Across both heterophilic and homophilic benchmarks, our enhanced \PPGs are competitive with---and often surpass---strong \MPG baselines, while preserving the scalability advantages.

\section{Related Work}
\label{sec:related}

\paragraph{Spectral graph neural networks.}
Early \emph{spectral} GNNs design graph convolutions as polynomial filters in the graph Fourier domain.
ChebNet~\cite{defferrard2016convolutional} popularized this line by approximating Laplacian spectral filters with truncated Chebyshev polynomials, avoiding explicit eigendecomposition.
Unlike \PPGs, ChebNet computes these polynomial filters \emph{on-the-fly} as part of the network's forward pass rather than as a preprocessing step.
LanczosNet~\cite{liao2019lanczosnet} further leverages Lanczos iterations to construct a \emph{global} low-rank Krylov approximation of the graph operator, yielding a feature-independent basis shared across channels.
In contrast to LanczosNet's shared basis, our approach constructs \emph{channel-specific} Krylov subspaces by using each feature channel as the Lanczos starting vector, producing per-channel tridiagonal matrix whose eigendecompositions yield channel-adaptive spectral bases.

\paragraph{Pre-propagation GNNs.}
\PPGs precompute multi-hop diffusions and train only a downstream predictor for scalability.
Representative models include SGC~\cite{wu2019SGC}, SIGN~\cite{frasca2020sign}, and adaptive hop aggregators such as GAMLP~\cite{zhang2022gamlp} and HOGA~\cite{deng2024hoga}.
Recent analysis suggests expressivity limitations of graph-augmented MLPs~\cite{chen2020graphaugmentedmlp}, motivating our stronger diffusion operators and lightweight re-coupling with representation learning.

\paragraph{Other scalable GNNs.}
Another line of scalable GNNs retains a propagation step but makes it inexpensive.
APPNP~\cite{gasteiger2018ppnp} performs a small number of power iterations that approximate Personalized PageRank, effectively enabling long-range information flow with controlled smoothing.
Correct-and-Smooth (C\&S)~\cite{huang2020cands} improves predictions by diffusing training residuals (\emph{correct}) and then smoothing outputs (\emph{smooth}), showing that simple propagation of model outputs can yield large gains when paired with a strong base predictor.

\paragraph{Label usage and prediction reuse.}
A related line of work leverages labels (or pseudo-labels) as additional signals.
UniMP~\cite{shi2020masked} injects training labels into the feature space and propagates them through message passing, unifying feature and label propagation.
SAGN with self-label enhancement (SLE) augments multi-hop feature models with label propagation and iterative pseudo-label refinement~\cite{sun2025scalable}.
Because label-derived features can introduce shortcut learning, these methods often pair propagation with regularization such as masking/dropout on label features or confidence filtering.
These approaches are conceptually close to reusing logits as node features; in contrast, our hidden-state re-propagation reuses task-adapted intermediate representations and empirically provides complementary information beyond class-probability signals.

\section{Conclusion}
We study the accuracy limitations of pre-propagation GNNs (\PPGs) and propose two scalable remedies.
First, we introduce robust diffusion operator families with better-conditioned diffusion bases than normalized-adjacency or random-walk diffusion.
Second, we propose few-shot hidden-state re-propagation (HRP), which occasionally diffuses intermediate representations to inject task-adapted signals without reverting to fully iterative message passing.
Across heterophilic and homophilic benchmarks, these components consistently improve \PPGs, narrowing the gap to competitive \MPGs and often matching or exceeding their accuracy, while maintaining training efficiency.
\section*{Acknowledgments}
This work was supported in part by ACE, one of the seven
centers in JUMP 2.0, a Semiconductor Research Corporation (SRC) program sponsored by DARPA and NSF Awards \#2212371 and \#2403135, and a research gift from Qualcomm. 
\section*{Impact Statement}
This work advances scalable graph representation learning by improving the accuracy of pre-propagation GNNs while preserving their efficiency benefits. As a general-purpose modeling and optimization technique, our methods may enable more effective use of graph-structured data in applications such as recommender systems, fraud detection, social and information network analysis, and scientific discovery, potentially reducing computational cost and energy consumption for large-scale graph learning.

At the same time, these applications can raise well-known ethical concerns. Improved performance on social or user-interaction graphs may amplify existing biases present in the data, impact fairness for underrepresented groups, or enable more effective profiling and targeted persuasion if deployed without safeguards. The proposed techniques do not introduce new data modalities or directly increase model access to sensitive attributes, but they also do not inherently prevent harmful downstream uses. Practitioners should therefore follow standard responsible-ML practices, including careful dataset governance, privacy protection, bias and fairness evaluation, and appropriate human oversight when deploying graph models in high-stakes settings.

We do not anticipate direct negative societal consequences beyond those already associated with learning on large graphs; nevertheless, we encourage future work on integrating fairness, privacy, and robustness constraints into scalable graph representation learning pipelines and on auditing the behavior of graph models under distribution shift and adversarial manipulation.
\bibliography{reference}

@string{DAC      = {Design Automation Conf. (DAC)}}

@string{ICLR     =  {International Conference on Learning Representations (ICLR)}}

@string{NIPS     = {Conf. on Neural Information Processing Systems ({NeurIPS})}}

@string{KDD      = {ACM SIGKDD Conf. on Knowledge Discovery \& Data Mining (KDD)}}

@string{MLSys    = {Machine Learning and Systems (MLSys)}}

@string{ICML     = {Int'l Conf. on Machine Learning (ICML)}}

@string{AAAI     = {AAAI Conf. on Artificial Intelligence (AAAI)}}

@string{WWW      = {Int'l World Wide Web Conf. (WWW)}}

@string{VLDB     = {Int'l Conf. on Very Large Data Bases (VLDB)}}

@string{IJCAI    = {Int'l Joint Conf. on Artificial Intelligence (IJCAI)}}

@article{kipf2016gcn,
  title="{Semi-Supervised Classification with Graph Convolutional Networks}",
  author={Kipf, Thomas N and Welling, Max},
  journal=ICLR,
  year={2017}
}

@article{velickovic2017gat,
  title="{Graph Attention Networks}",
  author={Veli{\v{c}}kovi{\'c}, Petar and Cucurull, Guillem and Casanova, Arantxa and Romero, Adriana and Lio, Pietro and Bengio, Yoshua},
  journal=ICLR,
  year={2018}
}

@article{hamilton2017sage,
  title="{Inductive Representation Learning on Large Graphs}",
  author={Hamilton, Will and Ying, Zhitao and Leskovec, Jure},
  journal=NIPS,
  year={2017}
}

@article{deng2024hoga,
  title="{Less is More: Hop-Wise Graph Attention for Scalable and Generalizable Learning on Circuits}",
  author={Deng, Chenhui and Yue, Zichao and Yu, Cunxi and Sarar, Gokce and Carey, Ryan and Jain, Rajeev and Zhang, Zhiru},
  journal=DAC,
  year={2024}
}

@article{wu2019SGC,
  title="{Simplifying Graph Convolutional Networks}",
  author={Wu, Felix and Souza, Amauri and Zhang, Tianyi and Fifty, Christopher and Yu, Tao and Weinberger, Kilian},
  journal=ICML,
  year={2019}
}

@article{frasca2020sign,
  title="{SIGN: Scalable Inception Graph Neural Networks}",
  author={Frasca, Fabrizio and Rossi, Emanuele and Eynard, Davide and Chamberlain, Ben and Bronstein, Michael and Monti, Federico},
  journal={arXiv preprint arXiv:2004.11198},
  year={2020}
}

@article{huang2020cands,
  title="{Combining Label Propagation and Simple Models Out-Performs Graph Neural Networks}",
  author={Huang, Qian and He, Horace and Singh, Abhay and Lim, Ser-Nam and Benson, Austin R},
  journal=ICLR,
  year={2021}
}

@article{chien2020gprgnn,
  title="{Adaptive Universal Generalized PageRank Graph Neural Network}",
  author={Chien, Eli and Peng, Jianhao and Li, Pan and Milenkovic, Olgica},
  journal=ICLR,
  year={2021}
}

@article{liao2022scara,
  title="{SCARA: Scalable Graph Neural Networks with Feature-Oriented Optimization}",
  author={Liao, Ningyi and Mo, Dingheng and Luo, Siqiang and Li, Xiang and Yin, Pengcheng},
  journal=VLDB,
  year={2022}
}

@article{bojchevski2020pprgo,
  title="{Scaling Graph Neural Networks with Approximate PageRank}",
  author={Bojchevski, Aleksandar and Gasteiger, Johannes and Perozzi, Bryan and Kapoor, Amol and Blais, Martin and R{\'o}zemberczki, Benedek and Lukasik, Michal and G{\"u}nnemann, Stephan},
  journal=KDD,
  year={2020}
}

@article{chen2020gbp,
  title="{Scalable Graph Neural Networks via Bidirectional Propagation}",
  author={Chen, Ming and Wei, Zhewei and Ding, Bolin and Li, Yaliang and Yuan, Ye and Du, Xiaoyong and Wen, Ji-Rong},
  journal=NIPS,
  year={2020}
}

@article{gasteiger2018ppnp,
  title="{Predict Then Propagate: Graph Neural Networks Meet Personalized PageRank}",
  author={Gasteiger, Johannes and Bojchevski, Aleksandar and G{\"u}nnemann, Stephan},
  journal=ICLR,
  year={2019}
}

@article{dong2021pta,
  title="{On the Equivalence of Decoupled Graph Convolution Network and Label Propagation}",
  author={Dong, Hande and Chen, Jiawei and Feng, Fuli and He, Xiangnan and Bi, Shuxian and Ding, Zhaolin and Cui, Peng},
  journal=WWW,
  year={2021}
}

@article{chen2020graphaugmentedmlp,
  title="{On Graph Neural Networks Versus Graph-Augmented MLPs}",
  author={Chen, Lei and Chen, Zhengdao and Bruna, Joan},
  journal=ICLR,
  year={2020}
}

@article{zhang2022gamlp,
  title="{Graph Attention Multi-Layer Perceptron}",
  author={Zhang, Wentao and Yin, Ziqi and Sheng, Zeang and Li, Yang and Ouyang, Wen and Li, Xiaosen and Tao, Yangyu and Yang, Zhi and Cui, Bin},
  journal=KDD,
  year={2022}
}

@article{zhu2020s2gc,
  title="{Simple Spectral Graph Convolution}",
  author={Zhu, Hao and Koniusz, Piotr},
  journal=ICLR,
  year={2021}
}

@article{gasteiger2019diffusion,
  title="{Diffusion Improves Graph Learning}",
  author={Gasteiger, Johannes and Wei{\ss}enberger, Stefan and G{\"u}nnemann, Stephan},
  journal=NIPS,
  year={2019}
}

@article{hu2020ogb,
  title="{Open Graph Benchmark: Datasets for Machine Learning on Graphs}",
  author={Hu, Weihua and Fey, Matthias and Zitnik, Marinka and Dong, Yuxiao and Ren, Hongyu and Liu, Bowen and Catasta, Michele and Leskovec, Jure},
  journal=NIPS,
  year={2020}
}

@article{gilmer2017mpgnn,
  title="{Neural Message Passing for Quantum Chemistry}",
  author={Gilmer, Justin and Schoenholz, Samuel S and Riley, Patrick F and Vinyals, Oriol and Dahl, George E},
  journal=ICML,
  year={2017}
}

@article{li2018oversmooth,
  title="{Deeper Insights Into Graph Convolutional Networks for Semi-Supervised Learning}",
  author={Li, Qimai and Han, Zhichao and Wu, Xiao-Ming},
  journal=AAAI,
  year={2018}
}

@article{lim2021hetero,
  title="{Large Scale Learning on Non-Homophilous Graphs: New Benchmarks and Strong Simple Methods}",
  author={Lim, Derek and Hohne, Felix and Li, Xiuyu and Huang, Sijia Linda and Gupta, Vaishnavi and Bhalerao, Omkar and Lim, Ser Nam},
  journal=NIPS,
  year={2021}
}

@article{nt2019lowpass,
  title="{Revisiting Graph Neural Networks: All We Have Is Low-Pass Filters}",
  author={Nt, Hoang and Maehara, Takanori},
  journal={arXiv preprint arXiv:1905.09550},
  year={2019}
}

@article{platonov2023critical,
  title={A critical look at the evaluation of GNNs under heterophily: Are we really making progress?},
  author={Platonov, Oleg and Kuznedelev, Denis and Diskin, Michael and Babenko, Artem and Prokhorenkova, Liudmila},
  journal=ICLR,
  year={2023}
}

@article{shchur1811pitfalls,
  title={Pitfalls of graph neural network evaluation},
  author={Shchur, Oleksandr and Mumme, Maximilian and Bojchevski, Aleksandar and G{\"u}nnemann, Stephan},
  journal={arXiv preprint arXiv:1811.05868},
  year={1811}
}

@article{yue2025graph,
  title={Graph Learning at Scale: Characterizing and Optimizing Pre-Propagation GNNs},
  author={Yue, Zichao and Deng, Chenhui and Zhang, Zhiru},
  journal=MLSys,
  year={2025}
}

@article{bo2021beyond,
  title={Beyond low-frequency information in graph convolutional networks},
  author={Bo, Deyu and Wang, Xiao and Shi, Chuan and Shen, Huawei},
  journal=AAAI,
  year={2021}
}

@article{defferrard2016convolutional,
  title={Convolutional neural networks on graphs with fast localized spectral filtering},
  author={Defferrard, Micha{\"e}l and Bresson, Xavier and Vandergheynst, Pierre},
  journal=NIPS,
  year={2016}
}

@article{wang2021bag,
  title={Bag of tricks for node classification with graph neural networks},
  author={Wang, Yangkun and Jin, Jiarui and Zhang, Weinan and Yu, Yong and Zhang, Zheng and Wipf, David},
  journal={arXiv preprint arXiv:2103.13355},
  year={2021}
}

@article{wang2022powerful,
  title={How powerful are spectral graph neural networks},
  author={Wang, Xiyuan and Zhang, Muhan},
  journal=ICML,
  year={2022}
}

@article{chen2020simple,
  title={Simple and deep graph convolutional networks},
  author={Chen, Ming and Wei, Zhewei and Huang, Zengfeng and Ding, Bolin and Li, Yaliang},
  journal=ICML,
  year={2020}
}

@article{zhu2020beyond,
  title={Beyond homophily in graph neural networks: Current limitations and effective designs},
  author={Zhu, Jiong and Yan, Yujun and Zhao, Lingxiao and Heimann, Mark and Akoglu, Leman and Koutra, Danai},
  journal=NIPS,
  year={2020}
}

@article{maurya2022simplifying,
  title={Simplifying approach to node classification in graph neural networks},
  author={Maurya, Sunil Kumar and Liu, Xin and Murata, Tsuyoshi},
  journal={Journal of Computational Science},
  volume={62},
  pages={101695},
  year={2022},
  publisher={Elsevier}
}

@article{li2022finding,
  title={Finding global homophily in graph neural networks when meeting heterophily},
  author={Li, Xiang and Zhu, Renyu and Cheng, Yao and Shan, Caihua and Luo, Siqiang and Li, Dongsheng and Qian, Weining},
  journal=ICML,
  year={2022}
}

@inproceedings{yan2022two,
  title={Two sides of the same coin: Heterophily and oversmoothing in graph convolutional neural networks},
  author={Yan, Yujun and Hashemi, Milad and Swersky, Kevin and Yang, Yaoqing and Koutra, Danai},
  booktitle={2022 IEEE International Conference on Data Mining (ICDM)},
  pages={1287--1292},
  year={2022},
  organization={IEEE}
}

@article{song2023ordered,
  title={Ordered gnn: Ordering message passing to deal with heterophily and over-smoothing},
  author={Song, Yunchong and Zhou, Chenghu and Wang, Xinbing and Lin, Zhouhan},
  journal=ICLR,
  year={2023}
}

@article{hua2022high,
  title={High-order pooling for graph neural networks with tensor decomposition},
  author={Hua, Chenqing and Rabusseau, Guillaume and Tang, Jian},
  journal=NIPS,
  year={2022}
}

@article{rusch2022gradient,
  title={Gradient gating for deep multi-rate learning on graphs},
  author={Rusch, T Konstantin and Chamberlain, Benjamin P and Mahoney, Michael W and Bronstein, Michael M and Mishra, Siddhartha},
  journal=ICLR,
  year={2023}
}

@inproceedings{rossi2024edge,
  title={Edge directionality improves learning on heterophilic graphs},
  author={Rossi, Emanuele and Charpentier, Bertrand and Di Giovanni, Francesco and Frasca, Fabrizio and G{\"u}nnemann, Stephan and Bronstein, Michael M},
  booktitle={Learning on graphs conference},
  pages={25--1},
  year={2024},
  organization={PMLR}
}

@article{liao2019lanczosnet,
  title={Lanczosnet: Multi-scale deep graph convolutional networks},
  author={Liao, Renjie and Zhao, Zhizhen and Urtasun, Raquel and Zemel, Richard S},
  journal=ICLR,
  year={2019}
}

@article{shi2020masked,
  title={Masked label prediction: Unified message passing model for semi-supervised classification},
  author={Shi, Yunsheng and Huang, Zhengjie and Feng, Shikun and Zhong, Hui and Wang, Wenjin and Sun, Yu},
  journal=IJCAI,
  year={2021}
}

@article{sun2025scalable,
  title={Scalable and adaptive graph neural networks with self-label-enhanced training},
  author={Sun, Chuxiong and Hu, Jie and Gu, Hongming and Chen, Jinpeng and Liang, Wei and Yang, Mingchuan},
  journal={Pattern Recognition},
  volume={160},
  pages={111210},
  year={2025},
  publisher={Elsevier}
}
\bibliographystyle{icml2026}

\newpage
\appendix
\onecolumn

\section{Dataset details}
\label{app:dataset}
\begin{table*}[t]
  \centering
  \captionsetup{width=\textwidth}
  \caption{Dataset statistics. $N$ and $E$ denote the number of nodes and edges (undirected edges counted once; directed graphs count directed edges).
  Split: ``fixed'' denotes the standard public split; ``random'' denotes averages over multiple random splits.}
  \vspace{-6pt}
  \label{tab:dataset_stats}
  \setlength{\tabcolsep}{6pt}
  \renewcommand{\arraystretch}{0.95}

  \begin{tabular}{lrrrrrrc}
    \toprule
    Dataset & $N$ & $E$ & Feat.\ dim & \#Classes & Train/Val/Test & Split & Type \\
    \midrule
    \texttt{roman-empire}   & $22,662$    & $32,927$     & $300$   & $18$ & 50/25/25 & fixed & Hetero \\
    \texttt{amazon-ratings} & $24,492$    & $93,050$     & $300$   & $5$  & 50/25/25 & fixed & Hetero \\
    \texttt{minesweeper}    & $10,000$    & $39,402$     & $7$     & $2$  & 50/25/25 & fixed & Hetero \\
    \texttt{tolokers}       & $11,758$    & $519,000$    & $10$    & $2$  & 50/25/25 & fixed & Hetero \\
    \texttt{questions}      & $48,921$    & $153,540$    & $301$   & $2$  & 50/25/25 & fixed & Hetero \\
    \texttt{pokec}          & $1,632,803$ & $30,622,564$ & $65$    & $2$  & 50/25/25 & fixed & Hetero \\
    \midrule
    \texttt{amazon-photo}    & $7,650$   & $119,081$   & $745$   & $8$  & 60/20/20 & random & Homo \\
    \texttt{amazon-computer} & $13,752$  & $245,861$   & $767$   & $10$ & 60/20/20 & random & Homo \\
    \texttt{coauthor-cs}     & $18,333$  & $81,894$    & $6,805$ & $15$ & 60/20/20 & random & Homo \\
    \texttt{coauthor-physics}& $34,493$  & $247,962$   & $8,415$ & $5$  & 60/20/20 & random & Homo \\
    \texttt{wikics}          & $11,701$  & $216,123$   & $300$   & $10$ & 5/45/50 & fixed & Homo \\
    \texttt{ogbn-arxiv}      & $169,343$ & $1,166,243$ & $128$   & $40$ & 53.7/17.6/28.7 & fixed & Homo \\
    \bottomrule
  \end{tabular}
  \vspace{-10pt}
\end{table*}

The $12$ datasets used in our experiments are summarized in Table~\ref{tab:dataset_stats}.

\textbf{Split protocol.} We follow the standard split protocol for each dataset.
For the four homophilic datasets \texttt{\seqsplit{amazon-computer}}, \texttt{\seqsplit{amazon-photo}}, \texttt{\seqsplit{coauthor-cs}}, and \texttt{\seqsplit{coauthor-physics}}, we report mean $\pm$ standard deviation over random splits.
All other datasets use their fixed public splits.

\section{Hardware settings}
\label{app:hardware}
For the training efficiency study, we use a Linux server with an NVIDIA A100 GPU (80GB memory, CUDA 12.6).
Our training pipeline is built on the scalable \PPG framework of \citet{yue2025graph}.

\section{\MPG baseline details}
\label{app:gnn-baselines}
We list the \MPG baselines used in our experiments in Table~\ref{tab:baseline}.
\begin{table}[t]
  \centering
  \captionsetup{width=\columnwidth}
  \caption{Message-passing GNN baselines and the original papers that introduced them.}
  \label{tab:baseline}
  \setlength{\tabcolsep}{5pt}
  \renewcommand{\arraystretch}{0.95}
  
  \begin{tabular}{ll}
    \toprule
    Baseline & Original paper \\
    \midrule
    GCN & \cite{kipf2016gcn} \\
    GAT (incl.\ GAT-sep) & \cite{velickovic2017gat} \\
    GraphSAGE & \cite{hamilton2017sage} \\
    APPNP & \cite{gasteiger2018ppnp} \\
    GPRGNN & \cite{chien2020gprgnn} \\
    PPRGo & \cite{bojchevski2020pprgo} \\
    GCNII & \cite{chen2020simple} \\
    H2GCN & \cite{zhu2020beyond} \\
    FSGNN & \cite{maurya2022simplifying} \\
    GloGNN & \cite{li2022finding} \\
    GGCN & \cite{yan2022two} \\
    O-GNN & \cite{song2023ordered} \\
    tGNN (a.k.a.\ tGCN) & \cite{hua2022high} \\
    G$^2$-GNN & \cite{rusch2022gradient} \\
    DIR-GNN & \cite{rossi2024edge} \\
    LINKX & \cite{lim2021hetero} \\
    \bottomrule
  \end{tabular}
  \vspace{-10pt}
\end{table}
\begin{table*}[t]
  \centering
  \captionsetup{width=\textwidth}
  \caption{Hyperparameter tuning settings for baseline models without publicly available results on given datasets.}
  \vspace{-6pt}
  \label{tab:baseline_hparams}
  \setlength{\tabcolsep}{6pt}
  \renewcommand{\arraystretch}{0.95}
  \scriptsize
  \resizebox{\textwidth}{!}{%
  \begin{tabular}{lp{0.30\textwidth}p{0.60\textwidth}}
    \toprule
    Model & Fixed hyperparameters & Tuned hyperparameters (search space) \\
    \midrule
    GCNII
      & Hidden dim $512$; LR $0.001$; epochs $2000$
      & Layers $\{5, 10\}$; dropout $\{0.3, 0.5, 0.7\}$; $\alpha \in \{0.3, 0.5, 0.7\}$; $\theta \in \{0.5, 1.0\}$ \\
    GGCN
      & Hidden dim $512$; LR $0.001$; epochs $2000$
      & Layers $\{5, 10\}$; dropout $\{0.3, 0.5, 0.7\}$; decay rate $\eta \in \{0.5, 1.0, 1.5\}$; exponent $\{2, 3\}$ \\
    OrderedGNN
      & Hidden dim $512$; LR $0.001$; epochs $2000$
      & Layers $\{5, 10\}$; dropout $\{0.3, 0.5, 0.7\}$; chunk size $\{4, 16, 64\}$ \\
    tGCN
      & Hidden dim $512$; LR $0.001$; epochs $2000$
      & Layers $\{2, 3\}$; dropout $\{0.3, 0.5, 0.7\}$; rank $\{256, 512\}$ \\
    G$^2$-GNN
      & Hidden dim $512$; LR $0.001$; epochs $2000$
      & Layers $\{5, 10\}$; dropout $\{0.3, 0.5, 0.7\}$; exponent $p \in \{2, 3, 4\}$ \\
    DIR-GNN
      & Hidden dim $512$; LR $0.001$; epochs $2000$; $\alpha = 0.5$; GATConv + jumping knowledge (``max'')
      & Layers $\{3, 5\}$; dropout $\{0.3, 0.5, 0.7\}$ \\
    \bottomrule
  \end{tabular}%
  }
  \vspace{-10pt}
\end{table*}

We use the accuracy numbers reported in the original baseline papers and the benchmarking study of \citet{platonov2023critical}.
For baselines without publicly available results on a given dataset, we tune hyperparameters using the search spaces in Table~\ref{tab:baseline_hparams}.

\section{\PPG baseline details}
\label{app:ppgnnbaseline}
\begin{table*}[t]
\centering
\small
\setlength{\tabcolsep}{6pt}
\renewcommand{\arraystretch}{1.15}
\caption{Optuna hyperparameter search spaces for SIGN, HOGA, GAMLP, and GARNN (model-specific and shared optimization parameters).}
\label{tab:ppgmodel-search-space}
\begin{tabular}{p{0.14\linewidth} p{0.26\linewidth} p{0.52\linewidth}}
\hline
\textbf{Model} & \textbf{Hyperparameter} & \textbf{Search space (as implemented)} \\
\hline
\textbf{HOGA} &
\texttt{lr} &
\(\mathrm{LogUniform}(10^{-4}, 5\times10^{-2})\) \\
\textbf{HOGA} &
\texttt{weight\_decay} &
\(\mathrm{LogUniform}(10^{-7}, 5\times10^{-3})\) \\
\textbf{HOGA} &
\texttt{hidden\_channels} &
\(\mathrm{Categorical}\{128,192,256,320,384,448,512,640\}\) \\
\textbf{HOGA} &
\texttt{mlp\_hidden} &
\(\mathrm{Categorical}\{128,256,320,384,512,640\}\) \\
\textbf{HOGA} &
\texttt{mlp\_dropout} &
\(\mathrm{Uniform}(0.0, 0.6)\) \\
\textbf{HOGA} &
\texttt{dropout} &
\(\mathrm{Uniform}(0.0, 0.6)\) \\
\textbf{HOGA} &
\texttt{input\_dropout} &
\(\mathrm{Uniform}(0.0, 0.5)\) \\
\textbf{HOGA} &
\texttt{mlplayers} &
\(\mathrm{Int}[1,4]\) \\
\textbf{HOGA} &
\texttt{num\_layers} &
\(\mathrm{Int}[1,4]\) \\
\textbf{HOGA} &
\texttt{num\_heads} &
\(\mathrm{Categorical}\{1,2,4,8\}\) \\
\textbf{HOGA} &
\texttt{attn\_dropout} &
\(\mathrm{Uniform}(0.0, 0.5)\) \\
\textbf{HOGA} &
\texttt{use\_post\_res} &
\(\mathrm{Categorical}\{0,1\}\) \\
\hline
\textbf{GARNN} &
\texttt{lr} &
\(\mathrm{LogUniform}(10^{-4}, 5\times10^{-2})\) \\
\textbf{GARNN} &
\texttt{weight\_decay} &
\(\mathrm{LogUniform}(10^{-7}, 5\times10^{-3})\) \\
\textbf{GARNN} &
\texttt{hidden\_channels} &
\(\mathrm{Categorical}\{128,192,256,320,384,448,512,640\}\) \\
\textbf{GARNN} &
\texttt{mlp\_hidden} &
\(\mathrm{Categorical}\{128,256,320,384,512,640\}\) \\
\textbf{GARNN} &
\texttt{mlp\_dropout} &
\(\mathrm{Uniform}(0.0, 0.6)\) \\
\textbf{GARNN} &
\texttt{dropout} &
\(\mathrm{Uniform}(0.0, 0.6)\) \\
\textbf{GARNN} &
\texttt{input\_dropout} &
\(\mathrm{Uniform}(0.0, 0.5)\) \\
\textbf{GARNN} &
\texttt{mlplayers} &
\(\mathrm{Int}[1,4]\) \\
\textbf{GARNN} &
\texttt{rnn\_type} &
\(\mathrm{Categorical}\{\texttt{LSTM},\texttt{GRU}\}\)\footnote{Overridable via \texttt{--search\_rnn\_types}.} \\
\textbf{GARNN} &
\texttt{rnn\_dim} &
\(\mathrm{Categorical}\{128,192,256,320,384,512\}\) \\
\textbf{GARNN} &
\texttt{num\_layers} &
\(\mathrm{Int}[1,3]\) \\
\textbf{GARNN} &
\texttt{rnn\_bidirectional} &
\(\mathrm{Categorical}\{\texttt{False},\texttt{True}\}\) \\
\textbf{GARNN} &
\texttt{rnn\_readout} &
\(\mathrm{Categorical}\{\texttt{rnn\_out},\texttt{hidden}\}\) \\
\textbf{GARNN} &
\texttt{use\_post\_res} &
\(\mathrm{Categorical}\{0,1\}\) \\
\textbf{GARNN} &
\texttt{use\_bn} &
\(\mathrm{Categorical}\{\texttt{False},\texttt{True}\}\) \\
\textbf{GARNN} &
\texttt{shared\_mlp} &
\(\mathrm{Categorical}\{\texttt{False},\texttt{True}\}\) \\
\hline
\textbf{GAMLP} &
\texttt{lr} &
\(\mathrm{LogUniform}(10^{-4}, 5\times10^{-2})\) \\
\textbf{GAMLP} &
\texttt{weight\_decay} &
\(\mathrm{LogUniform}(10^{-7}, 5\times10^{-3})\) \\
\textbf{GAMLP} &
\texttt{hidden\_channels} &
\(\mathrm{Categorical}\{128,192,256,320,384,448,512,640\}\) \\
\textbf{GAMLP} &
\texttt{mlp\_hidden} &
\(\mathrm{Categorical}\{128,256,320,384,512,640\}\) \\
\textbf{GAMLP} &
\texttt{mlp\_dropout} &
\(\mathrm{Uniform}(0.0, 0.6)\) \\
\textbf{GAMLP} &
\texttt{dropout} &
\(\mathrm{Uniform}(0.0, 0.6)\) \\
\textbf{GAMLP} &
\texttt{input\_dropout} &
\(\mathrm{Uniform}(0.0, 0.5)\) \\
\textbf{GAMLP} &
\texttt{mlplayers} &
\(\mathrm{Int}[1,4]\) \\
\textbf{GAMLP} &
\texttt{gamlp\_hidden} &
Deterministic: \(\texttt{gamlp\_hidden} \leftarrow \texttt{hidden\_channels}\) \\
\textbf{GAMLP} &
\texttt{gamlp\_alpha} &
\(\mathrm{Uniform}(0.1, 0.9)\) \\
\textbf{GAMLP} &
\texttt{gamlp\_n\_layers\_1} &
\(\mathrm{Int}[2,5]\) \\
\textbf{GAMLP} &
\texttt{gamlp\_n\_layers\_2} &
\(\mathrm{Int}[2,5]\) \\
\textbf{GAMLP} &
\texttt{gamlp\_input\_drop} &
\(\mathrm{Uniform}(0.0, 0.6)\) \\
\textbf{GAMLP} &
\texttt{gamlp\_att\_drop} &
\(\mathrm{Uniform}(0.0, 0.6)\) \\
\textbf{GAMLP} &
\texttt{gamlp\_act} &
\(\mathrm{Categorical}\{\texttt{relu},\texttt{leaky\_relu},\texttt{sigmoid}\}\) \\
\textbf{GAMLP} &
\texttt{gamlp\_pre\_process} &
\(\mathrm{Categorical}\{\texttt{False},\texttt{True}\}\) \\
\textbf{GAMLP} &
\texttt{gamlp\_residual} &
\(\mathrm{Categorical}\{\texttt{False},\texttt{True}\}\) \\
\textbf{GAMLP} &
\texttt{gamlp\_pre\_dropout} &
\(\mathrm{Categorical}\{\texttt{False},\texttt{True}\}\) \\
\textbf{GAMLP} &
\texttt{gamlp\_bns} &
\(\mathrm{Categorical}\{\texttt{False},\texttt{True}\}\) \\
\hline
\textbf{SIGN} &
\texttt{lr} &
\(\mathrm{LogUniform}(10^{-4}, 5\times10^{-2})\) \\
\textbf{SIGN} &
\texttt{weight\_decay} &
\(\mathrm{LogUniform}(10^{-7}, 5\times10^{-3})\) \\
\textbf{SIGN} &
\texttt{hidden\_channels} &
\(\mathrm{Categorical}\{128,256,384,512,640,768\}\) \\
\textbf{SIGN} &
\texttt{dropout} &
\(\mathrm{Uniform}(0.0, 0.7)\) \\
\textbf{SIGN} &
\texttt{input\_dropout} &
\(\mathrm{Uniform}(0.0, 0.5)\) \\
\textbf{SIGN} &
\texttt{num\_layers} &
\(\mathrm{Int}[2,5]\) \\
\hline
\end{tabular}
\end{table*}
To isolate the benefit of stronger diffusion operators \emph{under the same hop budget}, we fix the maximum hop budget to $15$ across methods and datasets.
We then use an automatic tuner to select the effective hop count, since the optimal number of hops can vary across operators and datasets due to oversmoothing and related effects; in practice, we typically find that fewer than $10$ hops suffice.
The hyperparameter ranges for each \PPG model are listed in Table~\ref{tab:ppgmodel-search-space}.
Unless otherwise stated, baseline \PPGs use two standard diffusions: normalized adjacency and random-walk diffusion.
To support HRP, we augment each \PPG backbone with an additional MLP layer before the final output projection, so that the hidden representations selected for re-propagation match the dimensionality of the original node features and can be blended directly.

\section{Spectrum-aware calibration for Jacobi bases.}
\label{app:calibration}
Let $L$ be the normalized graph Laplacian with spectrum $\lambda \in [0,2]$, and define the rescaled operator
\[
\tilde L \;=\; L - I ,
\]
whose eigenvalues $\tilde\lambda \in [-1,1]$.

To obtain a coarse estimate of the spectral density of $\tilde L$ without eigendecomposition, we employ a stochastic Chebyshev moment estimator.
We draw $R$ random probe vectors $z_r \sim \mathcal{N}(0,I)$ and compute
\[
v_0^{(r)} = z_r,\qquad
v_1^{(r)} = \tilde L z_r,\qquad
v_{k+1}^{(r)} = 2\tilde L v_k^{(r)} - v_{k-1}^{(r)} .
\]
The Chebyshev moments are estimated as
\[
m_k \;\approx\; \frac{1}{R}\sum_{r=1}^R \langle z_r, v_k^{(r)} \rangle
\;\approx\; \mathrm{Tr}\!\left(T_k(\tilde L)\right),
\qquad k=0,\dots,K_{\text{spec}} .
\]

From $\{m_k\}$, we reconstruct a coarse spectral density $\rho(\tilde\lambda)$ on $[-1,1]$ by matching moments on a fixed grid.
We then compute the low- and high-frequency spectral masses
\[
M_{\text{low}} = \int_{-1}^{0} \rho(\tilde\lambda)\, d\tilde\lambda,
\qquad
M_{\text{high}} = \int_{0}^{1} \rho(\tilde\lambda)\, d\tilde\lambda,
\]
and define the spectral imbalance
\[
\delta \;=\;
\frac{M_{\text{high}} - M_{\text{low}}}{M_{\text{high}} + M_{\text{low}}}
\;\in\;[-1,1].
\]

We use $\delta$ to select the parameters of the Jacobi polynomial basis.
Recall that Jacobi polynomials are orthogonal on $[-1,1]$ with respect to the weight
\[
w_{\alpha,\beta}(z) = (1-z)^{\alpha}(1+z)^{\beta}.
\]
We set
\[
\alpha = \gamma \min(-\delta,0),
\qquad
\beta  = \gamma \min(\delta,0),
\]
where $\gamma>0$ controls the strength of the bias.

This choice biases the polynomial basis toward regions of the spectrum with greater mass:
$\beta<0$ increases local resolution near $\tilde\lambda=-1$ (low Laplacian frequencies),
while $\alpha<0$ increases resolution near $\tilde\lambda=+1$ (high frequencies).
If $\delta\approx 0$, the method reduces to the Legendre basis $(\alpha=\beta=0)$.
The calibrated $(\alpha,\beta)$ are fixed during preprocessing and reused across all PP-GNN training stages.

\section{Optional auto-search details}
\label{app:auto-search}
\paragraph{Motivation.}
Two factors affect the effectiveness of HRP: (i) the stage-wise blending schedule, and (ii) the choice of checkpoint used to generate the re-propagated representations for the next stage.
Since re-propagation is applied outside the gradient path, the checkpoint with the best validation accuracy at stage $s$ is not always the one that yields the best re-propagated features for stage $s{+}1$.

\paragraph{Blending-weight tuning.}
When enabled, we use a small-budget covariance matrix adaptation evolution strategy (CMA-ES) to tune the per-hop blending weights $\{\alpha_s\}$ at each stage.
This search is performed with a small epoch budget and is disabled by default when computational resources are limited.

\paragraph{Checkpoint and operator selection.}
When enabled, we build a candidate set of epochs $\mathcal{E}_s$ that includes both early-epoch checkpoints and high-performing checkpoints.
We then select the best pair $(e,\Psi)$ using a two-stage screening procedure: a tight budget with a small hop count (e.g., $2$) for rapid filtering, followed by a larger budget for the finalist.

\paragraph{Spectral diversity score.}
To diversify candidates, we score how \emph{spectrally different} a hidden state is from the original input using per-channel spectral moments of the normalized Laplacian $L$.
For a feature matrix $X\in\mathbb{R}^{N\times d}$, let $X_c\in\mathbb{R}^{N}$ denote the $c$-th channel.
We define the (normalized) $k$-th moment signature
\begin{equation}
\mu_k(X_c) \;=\; \frac{\|L^k X_c\|_2^2}{\|X_c\|_2^2},
\qquad
\mathbf{m}(X_c) \;=\; \operatorname{norm}\!\bigl([\mu_0(X_c),\ldots,\mu_K(X_c)]\bigr),
\label{eq:method-spectral-moments}
\end{equation}
where $\operatorname{norm}(\cdot)$ normalizes the moment vector (e.g., to unit sum).
We compare two feature matrices by averaging a divergence between channel-wise signatures:
\begin{equation}
\operatorname{Dist}(X,Y)\;=\;\frac{1}{d}\sum_{c=1}^{d} D\!\left(\mathbf{m}(X_c),\mathbf{m}(Y_c)\right),
\label{eq:method-spectral-dist}
\end{equation}
with $D(\cdot,\cdot)$ a simple divergence (e.g., $\ell_1$ or Jensen--Shannon).
Each arm in the screening procedure corresponds to a candidate epoch and diffusion choice $(e,\Psi)$; we prioritize candidates that are both high-performing (train/validation) and spectrally diverse relative to the input features.
This step is executed only at stage boundaries; for very large graphs, we disable it by default.

\paragraph{Experimental setting and overhead.}
In our experiments, we first hand-tune HRP and enable the optional auto-search only when manual tuning does not already provide a substantive gain. In particular, the reported results on \texttt{physics} and \texttt{roman-empire} are obtained without CMA-ES; we also do not use CMA-ES on the two large graphs, \texttt{pokec} and \texttt{ogbn-arxiv}. On the remaining eight small datasets where CMA-ES is applied, we measure the overhead of the full optional tuning pipeline, including CMA-ES, checkpoint/operator screening, and moment-signature scoring, relative to fixed-configuration runs without auto-search. Averaged over the four \PPG backbones, the tuning overhead is $2.83\times$ on the homophilic datasets and $8.78\times$ on the heterophilic datasets. The additional memory cost is modest: the average increase in peak training memory is about 0.3 GB, with a maximum increase of 2 GB. These costs should therefore be interpreted as optional model-selection overhead rather than the intrinsic per-epoch cost of HRP.

\section{Hyperparameter Sensitivity of HRP}
\label{app:hrp_sensitivity}

To assess the hyperparameter sensitivity of HRP, we evaluate a fixed-configuration variant across all 12 datasets and four \PPG backbones. In this setting, HRP uses a fixed cosine blending schedule, without manual tuning of HRP hyperparameters or optional auto-search. Relative to the fully tuned configuration, this fixed variant incurs only a modest drop in average test performance: 0.38 points on heterophilic datasets, 0.31 points on homophilic datasets, and 0.34 points overall.

Importantly, even without tuning, HRP still provides a clear improvement over using robust diffusion operators alone, with an average gain of 0.51 points. Manual tuning or CMA-ES recovers an additional 0.34 points on average. These results suggest that the primary gain comes from the HRP mechanism itself, while optional tuning mainly serves as a lightweight convenience tool for extracting a smaller additional improvement.

\section{Diffusion Operator Choices}
\label{app:operators}
\makeatletter

\begin{table}[t]
\makeatother
\centering
\footnotesize
\setlength{\tabcolsep}{3pt}
\renewcommand{\arraystretch}{0.94}
\caption{Selected diffusion-operator-model pairs on different datasets used in Section~\ref{sec:eval}. Each row lists the preprocessing operator and the operator used for HRP for a given backbone--dataset pair. DAD denotes the symmetric normalized adjacency $D^{-1/2}AD^{-1/2}$, DA denotes the left-normalized diffusion $D^{-1}A$, Cheb and Leg denote Chebyshev and Legendre polynomial bases, Jac denotes the calibrated Jacobi basis, and Kry denotes the Lanczos/Krylov operator.}
\label{tab:operator-pairs}
\resizebox{\textwidth}{!}{%
\begin{tabular}{llcc@{\hspace{1.5em}}llcc}
\toprule
Model & Dataset & Pre-process & HRP & Model & Dataset & Pre-process & HRP \\
\midrule
HOGA & \texttt{amazon-computer} & Leg & Jac & HOGA & \texttt{ogbn-arxiv} & DAD & Jac \\
GARNN & \texttt{amazon-computer} & Leg & Jac & GARNN & \texttt{ogbn-arxiv} & DAD & Jac \\
GAMLP & \texttt{amazon-computer} & Leg & Jac & GAMLP & \texttt{ogbn-arxiv} & Kry & Jac \\
SIGN & \texttt{amazon-computer} & Leg & Jac & SIGN & \texttt{ogbn-arxiv} & DA & Jac \\
HOGA & \texttt{amazon-photo} & Leg & Jac & HOGA & \texttt{pokec} & Kry & Jac \\
GARNN & \texttt{amazon-photo} & Leg & Jac & GARNN & \texttt{pokec} & Kry & Jac \\
GAMLP & \texttt{amazon-photo} & Leg & Jac & GAMLP & \texttt{pokec} & DAD & Jac \\
SIGN & \texttt{amazon-photo} & Leg & Jac & SIGN & \texttt{pokec} & Kry & Jac \\
HOGA & \texttt{amazon-ratings} & DAD & Jac & HOGA & \texttt{questions} & DAD & Jac \\
GARNN & \texttt{amazon-ratings} & DA & Jac & GARNN & \texttt{questions} & Kry & Jac \\
GAMLP & \texttt{amazon-ratings} & DA & Jac & GAMLP & \texttt{questions} & DAD & Jac \\
SIGN & \texttt{amazon-ratings} & DA & Jac & SIGN & \texttt{questions} & DAD & Jac \\
HOGA & \texttt{coauthor-cs} & Leg & Kry & HOGA & \texttt{roman-empire} & Jac & Jac \\
GARNN & \texttt{coauthor-cs} & DAD & Jac & GARNN & \texttt{roman-empire} & Cheb & Jac \\
GAMLP & \texttt{coauthor-cs} & Leg & Jac & GAMLP & \texttt{roman-empire} & Leg & Jac \\
SIGN & \texttt{coauthor-cs} & Leg & Kry & SIGN & \texttt{roman-empire} & Jac & Jac \\
HOGA & \texttt{coauthor-physics} & DAD & Jac & HOGA & \texttt{tolokers} & Kry & Jac \\
GARNN & \texttt{coauthor-physics} & DAD & Jac & GARNN & \texttt{tolokers} & Kry & Jac \\
GAMLP & \texttt{coauthor-physics} & DAD & Jac & GAMLP & \texttt{tolokers} & DAD & Jac \\
SIGN & \texttt{coauthor-physics} & DAD & Jac & SIGN & \texttt{tolokers} & Jac & Jac \\
HOGA & \texttt{minesweeper} & Cheb & Jac & HOGA & \texttt{wikics} & DAD & Jac \\
GARNN & \texttt{minesweeper} & Cheb & Jac & GARNN & \texttt{wikics} & DA & Jac \\
GAMLP & \texttt{minesweeper} & Cheb & Jac & GAMLP & \texttt{wikics} & DA & Jac \\
SIGN & \texttt{minesweeper} & Jac & Jac & SIGN & \texttt{wikics} & DA & Jac \\
\bottomrule
\end{tabular}%
}
\makeatletter
\end{table}
\makeatother
Table~\ref{tab:operator-pairs} lists the diffusion-operator choices used for the results in Section~\ref{sec:eval}. For each dataset and \PPG backbone, we report the operator used for the initial preprocessing stage and the operator used during HRP. These choices define the fixed method configuration for each model--dataset pair.

The same operator choices are used in the non-CMA-ES experiments. In that setting, we disable the optional auto-search and use the fixed cosine blending schedule described in Appendix~\ref{app:hrp_sensitivity}; only the HRP blending and stage-selection procedure changes, while the preprocessing and HRP diffusion operators remain unchanged.

\section{Spectral Diagnostics of Krylov Preprocessing}
\label{app:ritz-diagnostics}
\begin{figure*}[t]
    \centering
    \begin{subfigure}[t]{0.48\textwidth}
        \centering
        \includegraphics[width=\linewidth]{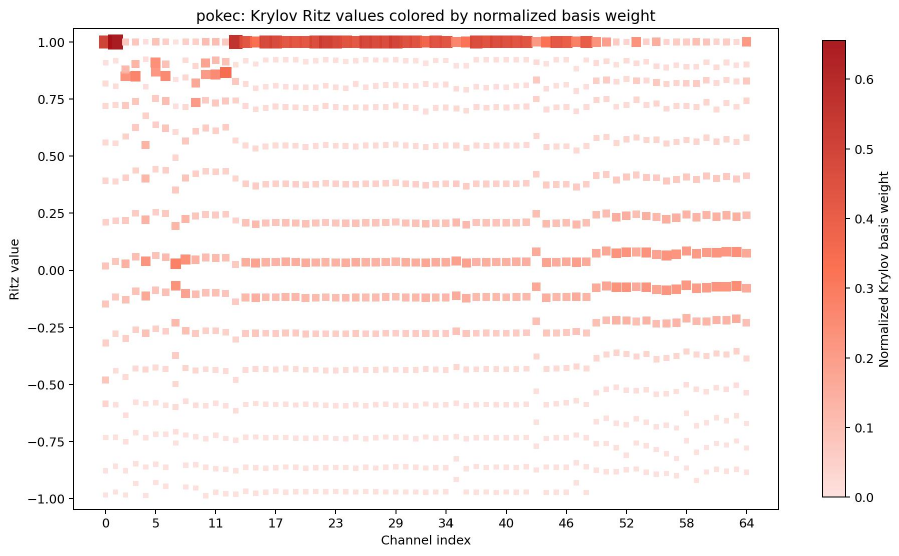}
        \caption{\texttt{pokec}: weighted Ritz-value map.}
        \label{fig:ritz-pokec}
    \end{subfigure}
    \hfill
    \begin{subfigure}[t]{0.48\textwidth}
        \centering
        \includegraphics[width=\linewidth]{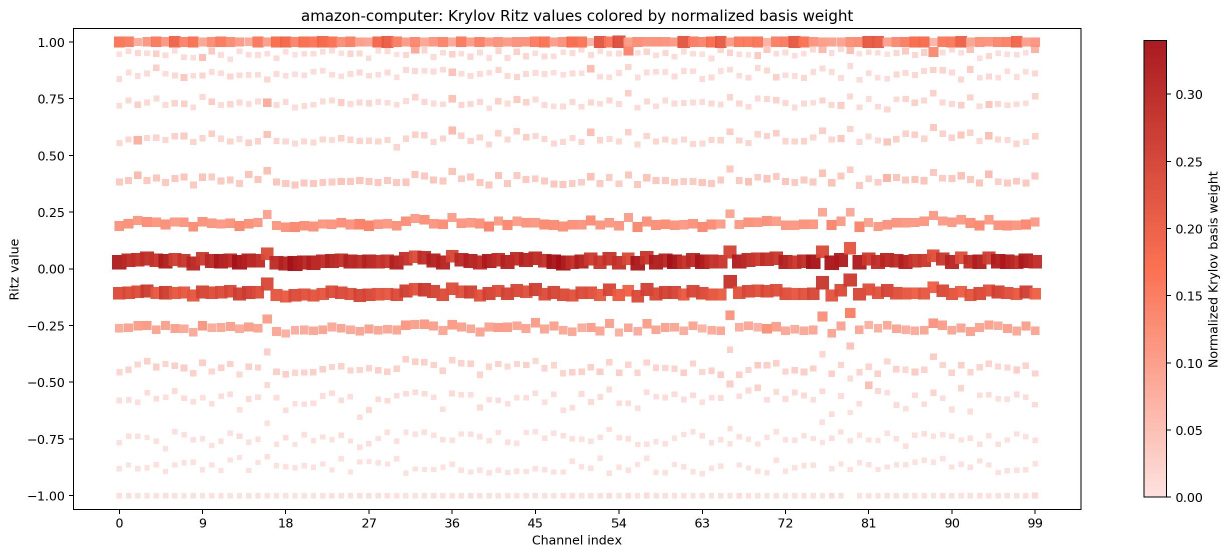}
        \caption{\texttt{amazon-computer}: weighted Ritz-value map.}
        \label{fig:ritz-amazon}
    \end{subfigure}

    \vspace{0.5em}

    \begin{subfigure}[t]{0.48\textwidth}
        \centering
        \includegraphics[width=\linewidth]{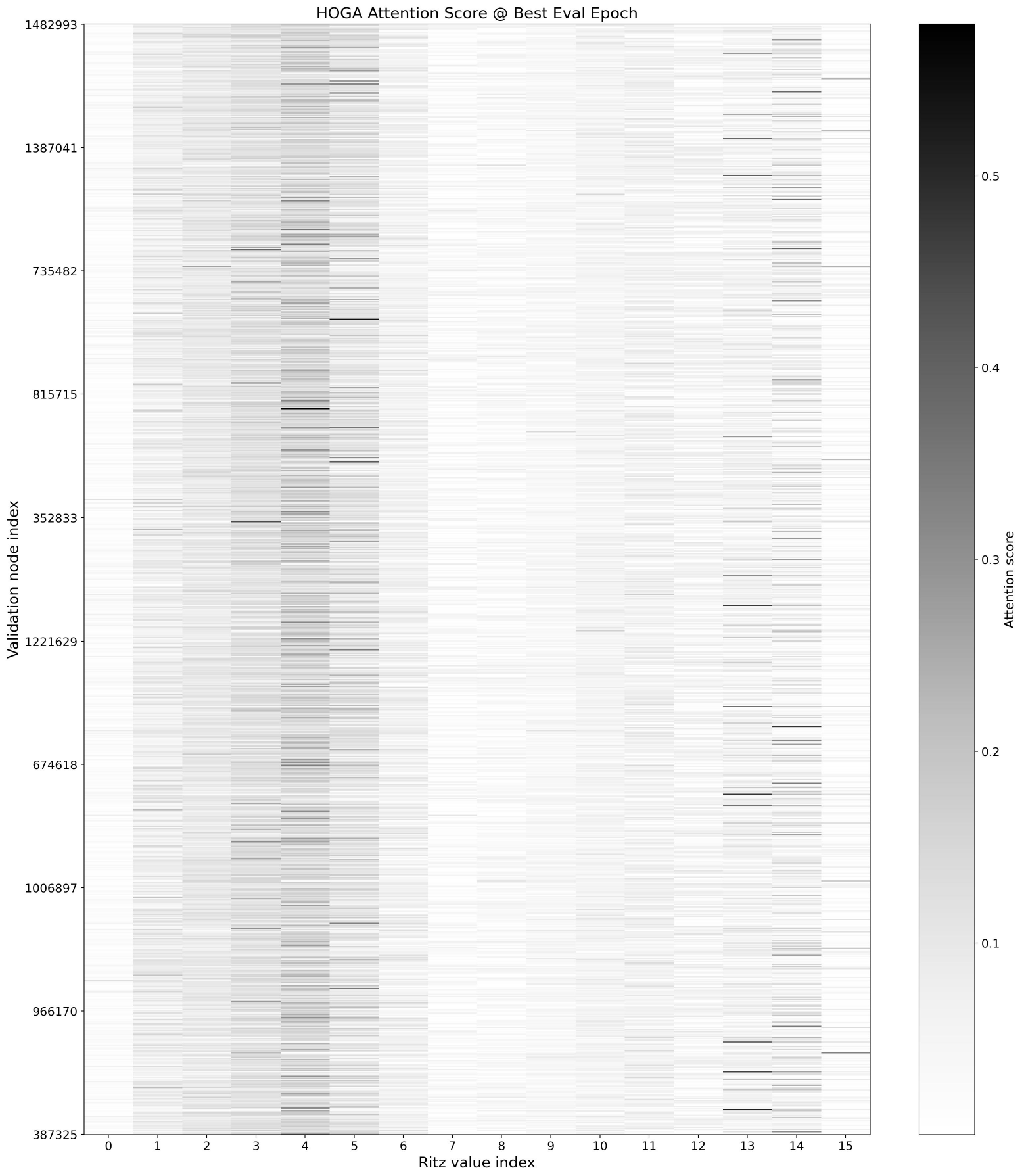}
        \caption{\texttt{pokec}: HOGA attention over Ritz indices.}
        \label{fig:atten-pokec}
    \end{subfigure}
    \hfill
    \begin{subfigure}[t]{0.48\textwidth}
        \centering
        \includegraphics[width=\linewidth]{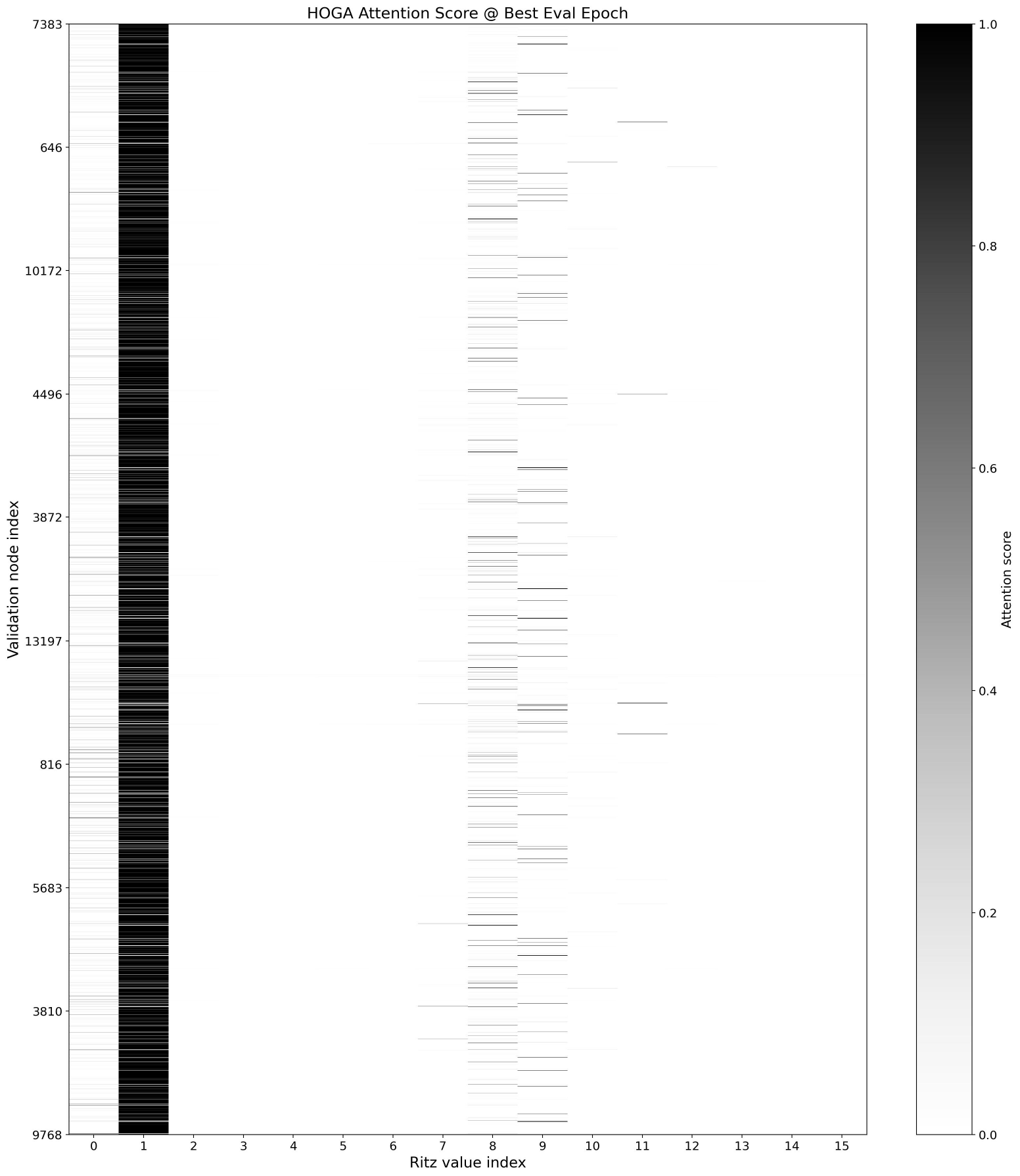}
        \caption{\texttt{amazon-computer}: HOGA attention over Ritz indices.}
        \label{fig:atten-amazon}
    \end{subfigure}

    \caption{
    Spectral diagnostics of Krylov preprocessing on a heterophilic graph, \texttt{pokec}, and a more homophilic graph, \texttt{amazon-computer}. 
    The top row shows weighted Ritz-value maps: each point corresponds to a channel-specific Ritz component, the vertical axis gives its Ritz value $\lambda_{c,i}$, and color indicates the normalized basis weight. 
    The bottom row shows HOGA attention scores over Ritz indices at the best validation epoch. 
    On \texttt{pokec}, preprocessing exposes more upper-spectrum components and the downstream attention is distributed across a broader range of Ritz indices. 
    On \texttt{amazon-computer}, attention is more concentrated on early Ritz indices, consistent with a stronger low-pass preference. 
    These plots characterize the spectrally meaningful inputs exposed to and emphasized by the \PPG backbone, rather than an exact spectral decomposition of the final nonlinear predictor.
    }
    \label{fig:ritz-diagnostics}
\end{figure*}
We provide qualitative diagnostics to examine what spectral components are exposed by the Krylov preprocessing and how the downstream \PPG backbone uses them. 
For each input channel $c$, the Lanczos procedure produces Ritz values $\lambda_{c,i}$ and corresponding Ritz components $z_{c,i}$, as described in Section~\ref{sec:method-filters}. 
Since $\tilde L = L-I$ has spectrum in $[-1,1]$, larger Ritz values correspond to higher-frequency components of the normalized Laplacian. 
Thus, the distribution of Ritz values provides a compact view of the spectral content made available to the downstream model.

We visualize two complementary quantities in Figure~\ref{fig:ritz-diagnostics}. 
First, the weighted Ritz-value maps show the components exposed by preprocessing. 
Each point corresponds to a channel-specific Ritz component, with vertical position given by its Ritz value $\lambda_{c,i}$ and color indicating its normalized basis weight. 
Second, the HOGA attention maps show how the downstream \PPG backbone uses the precomputed bank. 
We report attention scores at the best validation epoch, with rows corresponding to validation nodes and columns corresponding to Ritz indices.

On the heterophilic graph \texttt{pokec}, the weighted Ritz-value map places noticeable mass near the upper spectral range, indicating that Krylov preprocessing exposes substantial non-low-pass components. 
The corresponding HOGA attention map is also spread across a broader range of Ritz indices, suggesting that the downstream backbone does not rely only on early or low-pass components. 
In contrast, on the more homophilic \texttt{amazon-computer} graph, the exposed Ritz components are less concentrated at the high-frequency end, and the HOGA attention is more concentrated on early Ritz indices, which is consistent with a stronger low-pass preference.

These diagnostics should be interpreted at the level of the precomputed diffusion bank. 
The Krylov components have spectral meaning because each Ritz component is associated with a Ritz value, but the downstream \PPG backbone remains a nonlinear node-domain aggregator over the precomputed bank. 
Therefore, Figure~\ref{fig:ritz-diagnostics} is not an exact spectral decomposition of the final predictor; rather, it illustrates which spectrally meaningful inputs are exposed by preprocessing and emphasized by the \PPG backbone.

\end{document}